%% file: main.tex
\documentclass[10pt,twocolumn,letterpaper]{article}
\usepackage[pagenumbers]{wacv} 


\definecolor{wacvblue}{rgb}{0.21,0.49,0.74}
\usepackage[pagebackref,breaklinks,colorlinks,allcolors=wacvblue]{hyperref}
\def\wacvPaperID{2138} 
\def\confName{WACV}
\def\confYear{2026}

\title{ViP$^2$-CLIP: Visual-Perception Prompting with Unified Alignment for Zero-Shot Anomaly Detection}

\author{
  Ziteng Yang$^{1*}$\quad
  Jingzehua Xu$^{1*}$\quad
  Yanshu Li$^{2*}$\quad
  Zepeng Li$^{1}$\quad
  Yeqiang Wang$^{1}$\quad
  Xinghui Li$^{1\dagger}$\\[6pt]
  \normalsize
  $^{1}$Tsinghua University\quad
  $^{2}$Brown University
}
\usepackage{multirow}

\begin{document}

\maketitle

\input{sec/0_abstract}

\input{sec/1_intro}
\input{sec/2_formatting}

\input{sec/3_finalcopy}
\input{sec/4_method}
\input{sec/5_experiments}

\input{sec/6_conclusion}
\input{sec/7_Appeix}  
{
    \small
    \bibliographystyle{ieeenat_fullname}
    \bibliography{main}
}

\end{document}

%% file: sec/0_abstract.tex
\begin{abstract}
Zero‑shot anomaly detection (ZSAD) aims to detect anomalies without any target domain training samples, relying solely on external auxiliary data. Existing CLIP-based methods attempt to activate the model's ZSAD potential via handcrafted or static learnable prompts. The former incur high engineering costs and limited semantic coverage, whereas the latter apply identical descriptions across diverse anomaly types, thus fail to adapt to complex variations. Furthermore, since CLIP is originally pretrained on large-scale classification tasks, its anomaly segmentation quality is highly sensitive to the exact wording of class names, severely constraining prompting strategies that depend on class labels. To address these challenges, we introduce ViP$^{2}$-CLIP. The key insight of ViP$^{2}$-CLIP is a Visual‑Perception Prompting (ViP‑Prompt) mechanism, which fuses global and multi‑scale local visual context to adaptively generate fine‑grained textual prompts, eliminating manual templates and class-name priors. This design enables our model to focus on precise abnormal regions, making it particularly valuable when category labels are ambiguous or privacy-constrained. Extensive experiments on 15 industrial and medical benchmarks demonstrate that ViP$^{2}$-CLIP achieves state‑of‑the‑art performance and robust cross‑domain generalization.
\end{abstract}

%% file: sec/1_intro.tex
\section{Introduction}
\label{sec:intro}

\begin{figure}[htbp]
\centering
\includegraphics[width=0.9\linewidth]{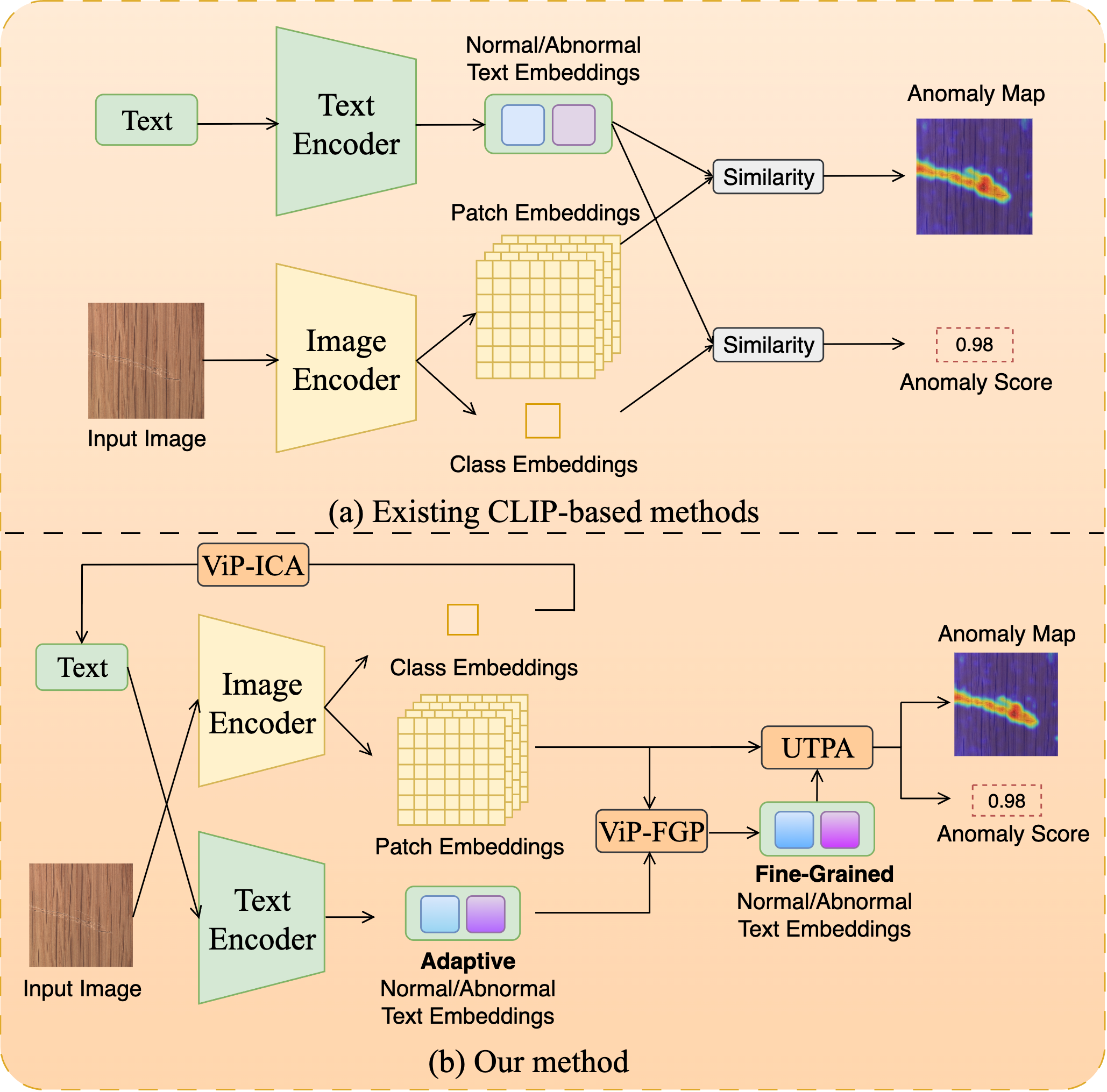}
\caption{Comparison between prior CLIP-based methods and ViP$^{2}$-CLIP. ViP$^{2}$-CLIP introduces ViP-Prompt to replace fixed class-name tokens with image-conditioned prompts that fuse global and local cues, and it first employs a unified patch-level alignment within training-based CLIP models.}
\label{fig:figure1}
\end{figure}

Large‑scale vision-language models (VLMs) such as CLIP \citep{radford2021learning} have shown impressive zero‑shot recognition capabilities by pre‑training on vast image-text pairs. Inspired by this success, CLIP has been widely adopted for Zero‐Shot Anomaly Detection (ZSAD) \citep{jeong2023winclip}, which seeks to detect anomalies automatically without any target domain training samples, offering a practical solution for data‑scarce scenarios. Unlike conventional image classification, which focuses on foreground semantics, anomaly detection emphasizes irregular regions on object surfaces. Consequently, existing CLIP‑based ZSAD methods typically construct two sets of text prompts–`normal' and `anomalous', and align them with image features in a shared semantic space. As illustrated in \cref{fig:figure1} (a), text prompts are aligned with the global image feature to infer image-level anomaly score, while alignment with local patch features then generates pixel-level anomaly maps. 

Currently, state-of-the-art ZSAD methods focus on prompt engineering to leverage CLIP's generalization. Representative approaches such as WinCLIP \citep{jeong2023winclip}, AnomalyCLIP \citep{zhou2023anomalyclip}, and VCP-CLIP \citep{qu2024vcp} either craft semantically rich manual templates or learn prompt tokens to boost performance. However, manual templates demand extensive human effort and cannot guarantee comprehensive semantic coverage. Although some works \citep{zhou2023anomalyclip} introduce object-agnostic learnable prompts to capture generic normal and abnormal concepts, they apply the same description across diverse anomaly types, hindering adaptability to complex semantic variations. Moreover, methods like CLIP-AD \citep{chen2024clip}, AdaCLIP \citep{cao2024adaclip}, FILo \citep{gu2024filo}, KANOCLIP \citep{li2025kanoclip}, and AA-CLIP \citep{ma2025aa} rely heavily on class names as prompt priors. Yet we find that CLIP's segmentation quality remains acutely sensitive to semantically equivalent class-name variations, even after fine-tuning on auxiliary anomaly data (see Appendix A for detailed analysis). This likely stems from the limited scope of downstream adaptation, which cannot fully stabilize local alignment against label shifts. Such sensitivity constrains the class-name usage in prompts, highlighting the urgent need for robust, class-name agnostic but adaptive prompting strategies in ZSAD.

To this end, we propose a Visual‑Perception Prompting (ViP‑Prompt) mechanism composed of an Image‑Conditioned Adapter (ICA) and a Fine‑Grained Perception Module (FGP). ICA adaptively injects global visual context into the prompts' embedding space, while FGP integrates multi‑scale patch features, enabling the prompts to capture fine‑grained irregularities. By replacing fixed class-name tokens with image-conditioned prompts that fuse global and local cues, ViP-Prompt significantly enhances prompts' robustness and reinforces cross-modal alignment across diverse semantic scenarios. Building upon that, we further introduce ViP$^2$-CLIP, a ZSAD model that integrates visual-perception prompts with a unified alignment mechanism; its overall architecture is depicted in \cref{fig:figure1} (b). ViP$^{2}$-CLIP first employs ViP-Prompt to condition text prompts on both global and local visual cues, fully leveraging CLIP's ZSAD performance; it then adopts a Unified Text-Patch Alignment (UTPA) strategy to align prompt tokens with multi-scale patch features through a single optimization, yielding superior precision in both detection and localization.

Our key contributions are summarized as follows:
\begin{itemize}[leftmargin=*]
\item We propose ViP-Prompt, which fuses global and local visual features to generate image-conditioned prompts, fully exploiting CLIP's potential for ZSAD. This design eliminates the need for handcrafted templates or class-name priors, delivers stronger generalization and robustness, making it particularly valuable when category labels are ambiguous or privacy‑constrained.
\item On top of that, we present ViP$^{2}$-CLIP, which integrates a simple but effective strategy-UTPA, that for the first time, introduces a consistency-alignment strategy within training-based CLIP models, enhancing CLIP's performance in both anomaly detection and localization.
\item Comprehensive experiments on 15 industrial and medical datasets demonstrate that ViP$^{2}$-CLIP consistently achieves superior ZSAD performance.
\end{itemize}

\begin{figure*}[htbp]
\centering
\includegraphics[width=1.0\linewidth]{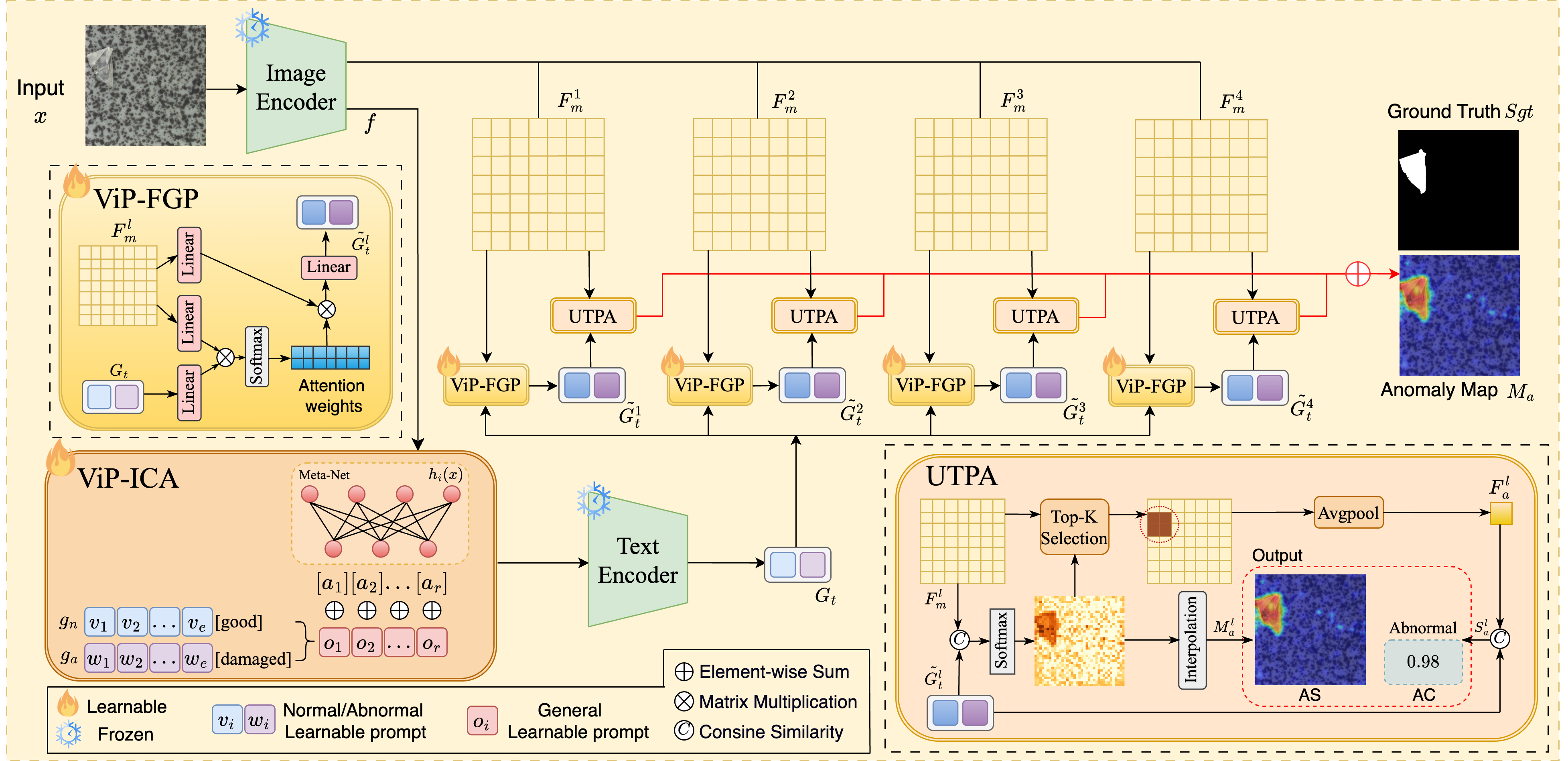}  
\caption{Framework of ViP$^{2}$-CLIP. ViP$^{2}$-CLIP first introduces ViP-Prompt to enhance cross-modal alignment: ViP-ICA injects global visual context into the prompts' embedding space, while ViP-FGP fuses local patch features to enhance the prompts' fine-grained perceptual capacity. Finally, the UTPA module performs unified alignment in multiple layers to jointly support image-level anomaly detection and pixel-level anomaly localization.}
\label{fig:figure2}
\end{figure*}

%% file: sec/2_formatting.tex
\section{Related Work}
\label{sec:related}

\paragraph{Zero-Shot Anomaly Detection (ZSAD)}  
ZSAD localizes abnormal regions without any target class samples during training. The advent of CLIP \citep{radford2021learning} has spurred notable advances in this task. WinCLIP \citep{jeong2023winclip} first ported CLIP to ZSAD with diverse handcrafted prompts. However, since CLIP is pretrained to align with general object semantics rather than abnormal patterns, its anomaly detection capability remains limited. To ease this burden, APRIL-GAN \citep{chen2023zero} and CLIP-AD \citep{chen2024clip} append shallow linear heads and fine-tune them on auxiliary datasets. AnomalyCLIP \citep{zhou2023anomalyclip} removes handcrafted templates via learnable, object-agnostic prompts, boosting cross-domain generalization, while AdaCLIP \citep{cao2024adaclip} employs hybrid prompts to align text and image features better. VCP-CLIP \citep{qu2024vcp} instead projects the image's global feature into a fixed handcrafted prompt space, which enhances segmentation performance. AA-CLIP \citep{ma2025aa} adopted a two-stage training strategy that separately optimizes text and image encoders to encourage greater divergence in the text embedding space for improved anomaly discrimination. AF-CLIP \citep{fang2025af} further integrated multi-scale feature aggregation to improve detection performance.

\paragraph{Prompt Learning}  
Prompt learning was initially proposed in natural language processing to enhance the adaptability of pre-trained models \citep{ouyang2022training, touvron2023llama} to diverse downstream tasks. CoOp \citep{radford2021learning} first brought this paradigm to the vision domain by inserting trainable tokens into text inputs, allowing CLIP to adapt to specific tasks without fine‑tuning. Static prompts, however, soon proved brittle on unseen classes. To enhance transferability, CoCoOp \citep{zhou2022conditional} and DenseCLIP \citep{rao2022denseclip} generate image-conditioned prompts that respond to visual context. Recent works \citep{yao2023visual, yao2024tcp} enforce prompt generalization by regularizing learnable tokens toward handcrafted templates, improving unseen class recognition. Works as MaPL \citep{khattak2023maple}, PromptSRC \citep{khattak2023self} and MMRL \citep{guo2025mmrl} push prompting into both modalities, jointly tuning image and text space for more substantial cross-modal alignment.

%% file: sec/3_finalcopy.tex
\section{Preliminary}

CLIP consists of a text encoder $T(\cdot)$ and a visual encoder $F(\cdot)$, both implemented as mainstream multi-layer networks. Leveraging contrastive learning on large-scale image-text pairs, CLIP achieves superior zero-shot recognition. Given a class name $c$, we combine it with a text prompt template $G$ (e.g., `A photo of a [cls]', where [cls] represents the category class-name). The resulting text is then fed into the text encoder $T(\cdot)$ to obtain the prompt embedding $g_{c} = T\bigl(G(c)\bigr) \in \mathbb{R}^{D}$. For an input image $x_{i}$, the visual encoder generates its global visual embedding $f_{i} \in \mathbb{R}^{D}$ and the local patch embeddings $f_{i}^{m} \in \mathbb{R}^{H\times W\times D}$. Specifically, given a category set $C$, CLIP computes the probability of image $x_{i}$ belonging to $c$ as follows:
\begin{equation}
{ p\bigl(y=c\mid{x_i}\bigr)=P(g_{c}, f_{i})
      = \frac{\exp\!\bigl(\langle g_{c},f_{i}\rangle/\tau\bigr)}
             {\sum_{c'\in C}\exp\!\bigl(\langle g_{c'},f_{i}\rangle/\tau\bigr)}
             },
\label{eq:eq1}
\end{equation}
where $\tau$ is a temperature hyperparameter and $\langle\cdot,\cdot\rangle$ denotes the cosine similarity.

Unlike conventional classification, ZSAD flags deviations from normality rather than assigning foreground semantics. Most approaches, therefore, instantiate two text prompts \cite{jeong2023winclip}, a normal prompt $g_{n}$ and an abnormal prompt $g_{a}$. At the image level, they compare the global image feature $f_{i}$ with both prompts and take $P(g_{a}, f_{i})$ as the anomaly score; while at the pixel level, for each location $(j,k)$, they extract the patch token $f_{i}^{m}(j,k)$, and compute the local normal score $S_{n}(j,k) = P\bigl(g_{n}, f_{i}^{m}(j,k)\bigr)$ and anomaly score $S_{a}(j,k) = P\bigl(g_{a}, f_{i}^{m}(j,k)\bigr)$, from which a pixel-wise anomaly map is generated.

%% file: sec/4_method.tex
\section{Method}

We propose ViP$^{2}$-CLIP, which leverages Visual-Perception Prompting (ViP-Prompt) to fully exploit CLIP's ZSAD capabilities. As illustrated in \cref{fig:figure2}, ViP$^{2}$-CLIP first introduces ViP-Prompt  (\cref{sec:prompt}) to fuse learnable normal and abnormal prompts with both global and multi-scale local visual features. This design enables prompts to adaptively capture the fine‑grained visual patterns of the inspected image, thereby improving cross-modal semantic alignment. Furthermore, we adopt a Unified Text-Patch Alignment (UTPA) scoring strategy (\cref{sec:utpa}), which optimizes alignment between prompt tokens and multi-scale patch features to support anomaly detection and localization jointly. By aggregating alignment signals across multiple layers, ViP$^{2}$-CLIP produces robust anomaly scores and fine-grained anomaly maps.

\subsection{Visual-Perception Prompt (ViP-Prompt)}
\label{sec:prompt}
We propose ViP-Prompt, an adaptive fine-grained prompting framework, which is built from an Image-Conditioned Adapter (ICA) and Fine-Grained Perception Module (FGP). By fusing learnable tokens with the global and local visual context, ViP-Prompt generates multi-level descriptions that flexibly track objects' visual patterns, thus markedly exploiting CLIP's ZSAD performance in diverse scenarios.
\paragraph{Image-Conditioned Adapter (ICA)}  
To eliminate dependence on handcrafted templates and class-name priors, we first define static learnable prompt templates for both normal and anomaly classes:
\begin{gather}
    g_n=[v_{1}][v_{2}]\dots[v_{e}]\;\text{good}\;[o_{1}]\dots[o_{r}], \\
    g_a=[w_{1}][w_{2}]\dots[w_{e}]\;\text{damaged}\;[o_{1}]\dots[o_{r}],
\end{gather}
here, $\{v_{i}\}, \{w_{i}\}\in\mathbb{R}^{C}$ denote the learnable normal and anomaly vector, respectively. $\{o_{i}\}\in\mathbb{R}^{C}$ is a generic learnable token that replaces the explicit class label. We adopt the adjectives `good' and `damaged' to guide the prompts to learn richer normal and anomalous semantics. By autonomously learning $g_{n}$ and $g_{a}$, the model can capture generic anomaly patterns across diverse objects.

To condition prompts on the target object and thus adapt to complex detection scenarios, we map the global visual embedding $f$ extracted by the image encoder through a lightweight Meta-Net $h_i(\cdot)$ (Linear–ReLU–Linear) into the text embedding space, denoted as $\{a_{i}\}_{i=1}^{r} = h_i(f)$, where $r$ is the number of mapped tokens and $a_{i}\in\mathbb{R}^{C}$, then fuse it with static learnable vector to generate dynamic prompt token. Thus, we define the prompts' structure as follows:
\begin{gather}
    g_n=[v_{1}][v_{2}]\dots[v_{e}]\;\text{good}\;[z_{1}]\dots[z_{r}], \\
    g_a=[w_{1}][w_{2}]\dots[w_{e}]\;\text{damaged}\;[z_{1}]\dots[z_{r}],
\end{gather}
where $z_{i} = o_{i} + a_{i}$. By projecting the image's global feature into the text embedding space, ICA adaptively generates normal and anomalous descriptors conditioned on the target object, eliminating the dependence on explicit class labels and enabling more robust cross‑modal alignment in privacy‑constrained settings.

\paragraph{Fine-Grained Perception Module (FGP)}
To further enhance the prompts' fine-grained perceptual capacity, we introduce an attention-based interaction module between the text prompts and multi-scale visual features. Specifically, we project both text embeddings and local patch embeddings into a shared $C$-dimension space. Let $G_{t}\in\mathbb{R}^{2\times C}$ be the prompt embeddings from the text encoder, and $F_{m}^{l}\in\mathbb{R}^{HW\times D}$ be the smoothed local visual embeddings from the $l$-th layer of the visual encoder, we then apply three learnable linear mappings to produce query $Q_{t}=G_{t}W_{q}$, key $K_{m}^{l}=F_{m}^{l}W_{k}$, and value $V_{m}^{l}=F_{m}^{l}W_{v}$, where $W_{q}\in\mathbb{R}^{C\times C}$, $W_{k}, W_{v}\in\mathbb{R}^{D\times C}$, are learnable projection matrices. 

Then, we compute scaled dot-product attention via $\mathrm{Softmax}\bigl(Q_{t}K_{m}^{l\top}/\sqrt{C}\bigr)$, and apply these weights to the values $V_{m}^{l}$, yielding enhanced token embeddings. Finally, a linear projection $W_{o}\in\mathbb{R}^{C\times D}$ transforms these embeddings into the layer-specific prompts $\tilde{G_{t}^{l}}$, which integrate fine-grained local features:
\begin{equation}
    \tilde{G_{t}^{l}} = \bigl(\mathrm{Softmax}(Q_{t} K_{m}^{l\top}/\sqrt{C})\,V_{m}^{l}\bigr)\,W_{o},
\end{equation}
here, $\tilde{G_{t}^{l}}\in\mathbb{R}^{2\times D}$ comprises the fine-grained normal prompt $\tilde{g_{n}}$ and anomalous prompt $\tilde{g_{a}}$.  
\begin{figure}[tp!]
\centering
\includegraphics[width=0.9\linewidth]{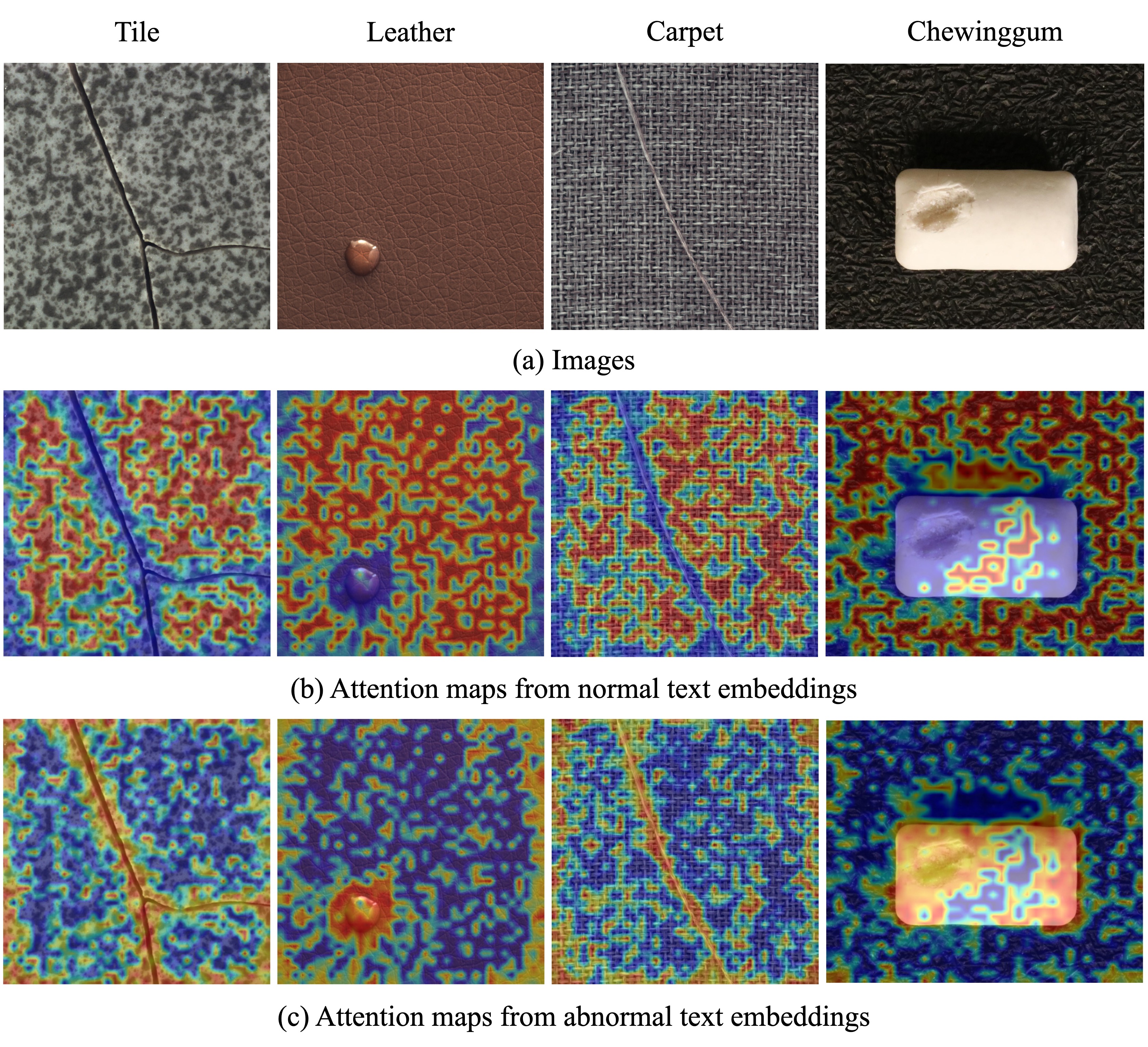} 
\caption{Visualization results of the attention maps from different prompts in our FGP module.}
\label{fig:figure3}
\end{figure}

To assess FGP's fine-grained sensitivity, \cref{fig:figure3} visualizes the attention maps of both normal and abnormal prompts across diverse images. Normal prompts activate on regular textures and structures of the image, while anomalous prompts concentrate on defects and irregular regions, indicating high sensitivity to anomalous patterns. These results demonstrate that FGP effectively refines prompt embeddings by incorporating localized visual cues, thereby enhancing cross-modal alignment.

\subsection{Unified Text-Patch Alignment (UTPA)}
\label{sec:utpa}
Recent CLIP-based ZSAD models typically adopt a dual-branch alignment scheme \cite{chen2024clip, zhou2023anomalyclip, chen2023zero}: one branch aligns the text prompt with global visual embedding to score anomalies at the image level, while the other aligns it with local patch embeddings for pixel-level anomaly maps. However, because these branches operate in separate feature subspaces, it hinders the simultaneous attainment of high-precision anomaly detection and localization.

To address this limitation, we introduce the UTPA strategy, which treats anomaly detection and localization as a unified alignment task between text embeddings and local patch embeddings. Concretely, to obtain the pixel-level anomaly map, we follow the standard practice: for the $l$-th layer feature map $F_{m}^l$, the patch token at spatial location $(j,k)$, denoted as $f_{j,k}^{l}\in\mathbb{R}^{D}$, receives its anomaly score by computing the similarity $P\bigl(\tilde{g_{a}},\,f_{j,k}^{l}\bigr)$, as defined in \cref{eq:eq1}.

Next, we upsample the anomaly scores of all patch tokens to produce the anomaly map $M_{a}^{l}$: 
\begin{equation}
    M_{a}^{l}
    =
    \mathrm{Up}\bigl(P(\tilde{g_{a}},\,F_m^{l})\bigr),
\end{equation}
where $Up(\cdot)$ denotes channel‐wise upsampling. 

Unlike prior works \citep{chen2024clip, zhou2023anomalyclip, chen2023zero} that rely on global alignment, we contend that an image's abnormality is driven by its most defective regions. Consequently, our image-level score is still derived from local patch features. For the $l$-th layer feature map $F_{m}^l$, we select the Top-k patch tokens with the highest $P\bigl(\tilde{g_{a}},\,f_{j,k}^{l}\bigr)$ among all patches. The image's representative anomaly feature $F_a^l$ is then obtained via global average pooling over this subset:
\begin{equation}
    F_{a}^{l} = \frac{1}{k}\sum_{(j,k)\in\mathcal{T}_{l}} f^{l}_{j,k},
\end{equation}
where \(\mathcal{T}_{l}\) indexes the chosen Top-k patches. The Top-k pooling directs the model's attention to the most suspicious patches, ensuring that the subsequent anomaly score faithfully reflects the severity of localized defects. Consequently, the image-level anomaly score $S_a^l$ at the $l$-th layer is obtained by computing $S_{a}^{l}=P\bigl(\tilde{g_{a}},\,F_{a}^{l}\bigr)$. By enforcing a unified text-patch alignment, UTPA effectively overcomes the optimization conflict of traditional dual-branch schemes, yielding more balanced ZSAD performance, as further validated in \cref{sec:ablation}.

\input{sec/table/table1}
\input{sec/table/table2}

\subsection{Training and Inference}

During training, ViP$^{2}$-CLIP is optimized by minimizing a global loss $L_{\mathrm{global}}$ and a local loss $L_{\mathrm{local}}$ \citep{zhou2023anomalyclip}: 
\begin{equation}
    L_{\mathrm{total}}=L_{\mathrm{global}}+\lambda \sum_{l=1}^{N} L_{\mathrm{local}}^{M^{l}},
\end{equation}
here, $\lambda$ is a hyperparameter to balance the global and local losses; $N$ represents the number of intermediate layers. $L_{\mathrm{global}}$ is the cross‐entropy loss matching the cosine similarity between text embeddings $\tilde{g_{n}}/\tilde{g_{a}}$ and the representative anomalous visual feature $F_{a}^{l}$; $L_{\mathrm{local}}$ is computed as the sum of a Focal loss \cite{lin2017focal} and a Dice loss \cite{li2019dice} to optimize local alignment jointly: 
\begin{equation}
    \begin{split}
    L_{\mathrm{local}}
    = \mathrm{Focal}\bigl(\mathrm{Up}([M_{n}^{l},M_{a}^{l}]),S_{\mathrm{gt}}\bigr) \\
    + \mathrm{Dice}\bigl(\mathrm{Up}(M_{n}^{l}),\,I - S_{\mathrm{gt}}\bigr) \\
    + \mathrm{Dice}\bigl(\mathrm{Up}(M_{a}^{l}),\,S_{\mathrm{gt}}\bigr),
    \end{split}
\end{equation}
where $[\cdot,\cdot]$ denotes channel‐wise concatenation; for each layer $l$, $M_{n}^{l}$ and $M_{a}^{l}$ are the normal and anomaly score maps; $S_{\mathrm{gt}}$ is the ground‐truth mask; and $I$ is the all‐ones matrix. 

During inference, the final image‐level anomaly score is computed as:$\mathrm{Score}=\frac{1}{N}\sum_{l=1}^{N} S_{a}^{l}$. For pixel‐level prediction, we merge all intermediate maps $M_{n}^{l}$ and $M_{a}^{l}$, then obtain the anomaly map $\mathrm{Map}\in\mathbb{R}^{H\times W}$ as follows:
\begin{equation}
    \mathrm{Map}
    =
    G_{\sigma}\!\Biggl(\frac{1}{N}\sum_{l=1}^{N}\Bigl(\tfrac{1}{2}\bigl(I-\mathrm{Up}(M_{n}^{l})\bigr)
    +\tfrac{1}{2}\,\mathrm{Up}(M_{a}^{l})\Bigr)\Biggr)
\end{equation}
where $G_{\sigma}$ denotes Gaussian smoothing.

%% file: sec/table/table1.tex
\begin{table*}[]

\resizebox{\linewidth}{!}{
\begin{tabular}{cccccccccc}
\hline
\noalign{\hrule height 0.5mm}
\multirow{2}{*}{Task} &
  \multirow{2}{*}{Datasets} &
   \multirow{2}{*}{$|\mathcal{C}|$}  &
  CLIP &
  WinCLIP &
  APRIL-GAN &
  AnomalyCLIP &
  AdaCLIP &
  AA-CLIP &
  ViP$^{2}$-CLIP \\ \cline{4-10}
   &  &   & \textit{OpenCLIP}  & \textit{CVPR 2023} & \textit{CVPRw 2023}  & \textit{ICLR 2024} & \textit{ECCV 2024} & \textit{CVPR 2025} & - \\ \hline
\multirow{8}{*}{\begin{tabular}[c]{@{}c@{}}Image-level\\ (AUROC, AP, F1)\end{tabular}} &
  MVTec AD &
  15 &
  (66.5, 82.6, 86.7) &
  (\textcolor{red}{91.8}, \textcolor{red}{96.5}, \textcolor{red}{92.7})$^{*}$ &
  (86.1, 93.5, 90.4) &
  (\textcolor{blue}{91.6}, \textcolor{blue}{96.4}, \textcolor{red}{92.7}) &
  (90.1, 95.6, \textcolor{blue}{92.3}) &
  (89.1, 94.7, 90.0) &
  (91.2, 96.0, 92.0) \\
 &
  VisA &
  12 &
  (60.2, 66.2, 73.0) &
  (78.1, 81.2, 78.2)$^{*}$ &
  (77.5, 80.9, 78.7) &
  (82.0, 85.3, 80.4) &
  (\textcolor{blue}{87.2}, \textcolor{blue}{89.7}, \textcolor{blue}{83.5}) &
  (78.9, 82.3, 78.8) &
  (\textcolor{red}{88.5}, \textcolor{red}{90.4}, \textcolor{red}{84.8}) \\
 &
  MPDD &
  6 &
  (58.7, 69.7, 76.0) &
  (61.5, 69.2, 77.5) &
  (76.8, 83.0, 81.0) &
  (\textcolor{blue}{77.5}, \textcolor{blue}{82.5}, 80.4) &
  (74.8, 78.6, \textcolor{red}{83.3}) &
  (56.9, 66.5, 74.5) &
  (\textcolor{red}{79.7}, \textcolor{red}{84.5}, \textcolor{blue}{82.4}) \\
 &
  KSDD &
  1 &
  (74.3, 55.3, 57.8) &
  (92.4, 82.9, 77.7) &
  (96.5, 91.2, 85.4) &
  (\textcolor{blue}{97.8}, \textcolor{blue}{94.2}, \textcolor{blue}{89.7}) &
  (97.2, 92.6, 89.5) &
  (96.0, 89.2, 84.7) &
  (\textcolor{red}{98.1}, \textcolor{red}{95.8}, \textcolor{red}{93.2}) \\
 &
  BTAD &
  3 &
  (25.7, 49.8, 66.0) &
  (68.2, 70.9, 67.8) &
  (73.7, 69.7, 68.2) &
  (88.2, 88.2, 83.8) &
  (89.3, 96.5, 90.9) &
  (\textcolor{blue}{93.4}, \textcolor{blue}{97.6}, \textcolor{blue}{93.9})&
  (\textcolor{red}{95.0}, \textcolor{red}{98.4}, \textcolor{red}{94.7}) \\
 &
  DAGM &
  10 &
  (55.3, 43.2, 49.6) &
  (91.8, 79.5, 75.7) &
  (94.4, 83.9, 80.2) &
  (97.7, \textcolor{blue}{92.4}, 90.1) &
  (\textcolor{blue}{98.2}, 92.3, \textcolor{blue}{90.9}) &
  (95.3, 87.5, 84.5) &
  (\textcolor{red}{98.5}, \textcolor{red}{94.3}, \textcolor{red}{92.6}) \\
 &
  DTD-Synthetic &
  12 &
  (52.3, 77.8, 85.1) &
  (95.1, 97.7, 94.1) &
  (85.6, 94.0, 89.1) &
  (93.9, 97.2, 93.6) &
  (\textcolor{red}{96.3}, \textcolor{red}{98.1}, \textcolor{red}{95.5}) &
  (94.0, \textcolor{blue}{98.0}, \textcolor{blue}{94.3}) &
  (\textcolor{blue}{95.5}, \textcolor{red}{98.1}, \textcolor{blue}{94.3}) \\ \cline{2-10}
 &
  AVERAGE & -
   &
  (56.1, 63.5, 70.6) &
  (82.1, 82.1, 80.5) &
  (84.4, 85.2, 81.9) &
  (89.8, 90.9, 87.2) &
  (\textcolor{blue}{90.4}, \textcolor{blue}{91.9}, \textcolor{blue}{89.4}) &
  (86.7, 88.0, 86.5) &
  (\textcolor{red}{92.4}, \textcolor{red}{93.9},\textcolor{red}{90.6}) \\ \hline
\multirow{8}{*}{\begin{tabular}[c]{@{}c@{}}Pixel-level\\ (AUROC, PRO, F1)\end{tabular}} &
  MVTec AD &
  15 &
  (35.6, 10.6, 6.9) &
  (85.1, 64.6, 24.8)$^{*}$ &
  (87.6, 44.0, 43.3) &
  (\textcolor{blue}{91.1}, 81.4, 39.1) &
  (89.6, 37.8, \textcolor{blue}{45.1}) &
  (\textcolor{red}{91.5}, \textcolor{blue}{86.5}, \textcolor{red}{46.7}) &
  (90.5, \textcolor{red}{87.1}, 43.1) \\
 &
  VisA &
  12 &
  (43.6, 14.0, 1.5) &
  (79.6, 56.8, 9.0)$^{*}$ &
  (94.2, 86.6, 32.3) &
  (\textcolor{red}{95.5}, \textcolor{blue}{86.7}, 28.3) &
  (\textcolor{red}{95.5}, 56.8, \textcolor{red}{37.0}) &
  (94.7, 82.7, 29.4) &
  (\textcolor{blue}{95.4}, \textcolor{red}{92.2}, \textcolor{blue}{33.6}) \\
 &
  MPDD &
  6 &
  (56.2, 27.3, 8.2) &
  (71.2, 40.5, 15.4) &
  (94.3, 83.8, 31.3) &
  (\textcolor{blue}{96.5}, \textcolor{blue}{88.7}, \textcolor{blue}{34.2}) &
  (96.1, 60.3, 31.9) &
  (96.0, 86.6, 26.5) &
  (\textcolor{red}{97.2}, \textcolor{red}{92.6}, \textcolor{red}{35.9}) \\
 &
  KSDD &
  1 &
  (21.7, 2.6, 0.5) &
  (92.8, 70.3, 15.8) &
  (93.2, 84.1, 43.6) &
  (98.1, \textcolor{blue}{94.9}, \textcolor{red}{56.5}) &
  (98.4, 53.0, 52.8) &
  (\textcolor{red}{99.0}, 92.2, 48.1) &
  (\textcolor{blue}{98.5}, \textcolor{red}{96.2}, \textcolor{blue}{53.0}) \\
 &
  BTAD &
  3 &
  (41.2, 10.8, 6.8) &
  (72.7, 27.5, 18.5) &
  (89.3, 68.7, 40.6) &
  (\textcolor{blue}{94.2}, \textcolor{blue}{75.4}, 49.7) &
  (90.7, 22.3, \textcolor{blue}{51.4}) &
  (\textcolor{blue}{94.2}, 70.0, 45.1) &
  (\textcolor{red}{95.6}, \textcolor{red}{86.1}, \textcolor{red}{52.7}) \\
 &
  DAGM &
  10 &
  (49.6, 12.4, 2.4) &
  (87.6, 65.7, 12.7) &
  (82.4, 66.0, 37.4) &
  (\textcolor{blue}{95.6}, \textcolor{blue}{91.0}, 58.9) &
  (94.3, 42.5, \textcolor{blue}{59.6}) &
  (92.6, 79.7, 44.0) &
  (\textcolor{red}{97.5}, \textcolor{red}{95.2}, \textcolor{red}{61.2}) \\
 &
  DTD-Synthetic &
  12 &
  (43.8, 16.0, 3.7) &
  (79.5, 51.4, 16.1) &
  (95.2, 87.3, 67.4) &
  (97.9, \textcolor{blue}{92.0}, 62.2) &
  (\textcolor{blue}{98.5}, 75.0, \textcolor{red}{71.8}) &
  (97.4, 88.9, 58.0) &
  (\textcolor{red}{99.0}, \textcolor{red}{96.5}, \textcolor{blue}{67.5}) \\ \cline{2-10}
 &
  AVERAGE & -
   &
  (41.7, 13.4, 4.3) &
  (79.9, 52.6, 16.0) &
  (90.9, 74.4, 42.2) &
  (\textcolor{blue}{95.6}, \textcolor{blue}{87.2}, 47.0) &
  (94.7, 49.7, \textcolor{red}{49.9}) &
  (95.0, 83.8, 42.5) &
  (\textcolor{red}{96.2},\textcolor{red}{92.3}, \textcolor{blue}{49.6}) \\ \hline
  \noalign{\hrule height 0.5mm}
\end{tabular}
}
\caption{ZSAD performance comparison on industrial domain. \textsuperscript{*} denotes results taken from original papers. The best performance is shown in red, with the second-best highlighted in blue.}
\label{tab:table1}
\end{table*}

%% file: sec/table/table2.tex
\begin{table*}[ht]
\resizebox{\linewidth}{!}{
\begin{tabular}{cccccccccc}
\hline
\noalign{\hrule height 0.5mm}
\multirow{2}{*}{Task} &
  \multirow{2}{*}{Datasets} &
   \multirow{2}{*}{$|\mathcal{C}|$}  &
  CLIP &
  WinCLIP &
  APRIL-GAN &
  AnomalyCLIP &
  AdaCLIP &
  AA-CLIP &
  ViP$^{2}$-CLIP \\ \cline{4-10}
   &  &   & \textit{OpenCLIP}  & \textit{CVPR 2023} & \textit{CVPRw 2023}  & \textit{ICLR 2024} & \textit{ECCV 2024} & \textit{CVPR 2025} & - \\ \hline
\multirow{4}{*}{\begin{tabular}[c]{@{}c@{}}Image-level\\ (AUROC, AP, F1)\end{tabular}} &
  HeadCT & 1 & (67.8, 62.4, 70.9) & (81.8, 80.2, 78.9) & (89.1, 89.4, 82.1) & (93.0, 91.1, \textcolor{blue}{88.4}) & (\textcolor{blue}{94.0}, \textcolor{blue}{91.4}, \textcolor{red}{90.1}) & (88.1, 90.4, 80.6) & (\textcolor{red}{94.3}, \textcolor{red}{93.9}, 88.1) \\
 &
  BrainMRI & 1 & (72.2, 81.5, 76.5) & (86.6, 91.5, 84.1) & (89.4, 91.0, 88.2) & (90.0, 92.1, 86.5) & (\textcolor{blue}{94.3}, \textcolor{blue}{95.5}, \textcolor{blue}{92.2}) & (92.2, 94.5, 88.6) & (\textcolor{red}{95.3}, \textcolor{red}{96.7}, \textcolor{red}{92.3}) \\
 &
  Brain35H & 1 & (76.3, 77.7, 72.2) & (79.9, 82.2, 74.0) & (91.6, 92.1, 84.5) & (93.4, 93.8, 86.4) & (\textcolor{blue}{95.7}, \textcolor{blue}{95.8}, \textcolor{red}{91.1}) & (89.3, 90.6, 81.6) & (\textcolor{red}{95.8}, \textcolor{red}{96.0}, \textcolor{blue}{90.1}) \\ \cline{2-10}
 & 
  AVERAGE & - & (72.1, 73.9, 73.2) & (82.8, 84.6, 79.0) & (90.0, 90.8, 84.9) & (92.1, 92.3, 87.1) & (\textcolor{blue}{94.7}, \textcolor{blue}{94.2}, \textcolor{red}{91.1}) & (89.9, 91.8, 83.6) & (\textcolor{red}{95.1}, \textcolor{red}{95.5}, \textcolor{blue}{90.2}) \\
\hline
\multirow{6}{*}{\begin{tabular}[c]{@{}c@{}}Pixel-level\\ (AUROC, PRO, F1)\end{tabular}} 
& ISIC & 1 & (43.2, 7.6, 44.0) & (83.3, 55.1, 64.1) & (89.4, 77.2, 71.4) & (89.4, 78.4, 71.6) & (\textcolor{blue}{91.3}, 53.2, \textcolor{blue}{75.5}) & (\textcolor{red}{91.6}, \textcolor{red}{86.2}, \textcolor{red}{78.1}) & (90.3, \textcolor{blue}{82.3}, 73.7) \\
& CVC-ColonDB & 1 & (59.8, 30.4, 17.8) & (64.8, 28.4, 21.0) & (78.4, 64.6, 29.7) & (\textcolor{blue}{81.9}, \textcolor{blue}{71.2}, \textcolor{red}{37.5}) & (81.5, 64.3, 33.6) & (80.6, 62.1, 32.3) & (\textcolor{red}{82.5}, \textcolor{red}{74.8}, \textcolor{blue}{36.2}) \\
& CVC-ClinicDB & 1 & (63.1, 33.8, 23.4) & (70.3, 32.5, 27.2) & (80.5, 60.7, 38.7) & (82.9, \textcolor{blue}{68.1}, 42.4) & (83.9, 65.7, 42.3) & (\textcolor{blue}{85.6}, 66.3, \textcolor{blue}{44.3}) & (\textcolor{red}{86.3}, \textcolor{red}{72.1}, \textcolor{red}{44.9}) \\
& Endo & 1 & (59.7, 25.2, 28.9) & (68.2, 28.3, 32.9) & (81.9, 54.9, 44.8) & (84.2, 63.4, 50.3) & (\textcolor{red}{86.0}, 63.4, \textcolor{blue}{51.8}) & (\textcolor{red}{86.0}, \textcolor{red}{66.5}, \textcolor{red}{54.0}) & (\textcolor{blue}{84.5}, \textcolor{blue}{63.7}, 49.2) \\
& Kvasir & 1 & (58.0, 18.3, 29.3) & (69.7, 24.5, 35.9) & (75.0, 36.3, 40.0) & (79.0, 45.4, 46.2) & (\textcolor{blue}{81.6}, \textcolor{red}{49.1}, 47.1) & (81.5, \textcolor{blue}{48.4}, \textcolor{red}{49.2}) & (\textcolor{red}{81.9}, 47.8, \textcolor{blue}{47.3}) \\ \cline{2-10}
& AVERAGE & - & (56.8, 23.1, 28.7) & (71.3, 33.8, 36.2) & (81.0, 58.7, 44.9) & (83.5, 65.3, 49.6) & (\textcolor{blue}{84.9}, 59.1, 50.1) & (\textcolor{red}{85.1}, \textcolor{blue}{65.9}, \textcolor{red}{51.6}) & (\textcolor{red}{85.1}, \textcolor{red}{68.1}, \textcolor{blue}{50.3}) \\
\hline
\noalign{\hrule height 0.5mm}
\end{tabular}
}
\caption{ZSAD performance comparison on medical domain. The best performance is shown in red, with the second-best highlighted in blue. Note that the image-level medical AD datasets do not provide pixel-level segmentation annotations, so the pixel-level medical AD datasets are different from the image-level datasets.}
\label{tab:table2}
\end{table*}

%% file: sec/5_experiments.tex
\section{Experiments}

\subsection{Setup}

\paragraph{Datasets \& Baselines} 
To comprehensively evaluate ViP$^{2}$-CLIP across diverse application scenarios, we conduct extensive experiments on public benchmarks spanning industrial and medical domains. In the industrial domain, we use MVTec AD \citep{bergmann2019mvtec}, VisA \citep{zou2022spot}, MPDD \citep{jezek2021deep}, BTAD \citep{mishra2021vt}, KSDD \citep{tabernik2020segmentation}, DAGM \citep{wieler2007weakly}, and DTD-Synthetic \citep{aota2023zero}. For medical imaging, we include brain tumour detection datasets HeadCT \citep{salehi2021multiresolution}, BrainMRI \citep{salehi2021multiresolution}, and Br35H \citep{hamada2020br35h}, skin cancer detection dataset ISIC \citep{codella2018skin}, colon polyp detection datasets CVC-ClinicDB \citep{bernal2015wm}, CVC-ColonDB \citep{tajbakhsh2015automated}, Endo \citep{hicks2021endotect}, and Kvasir \citep{jha2019kvasir}. We compare ViP$^{2}$-CLIP against six leading ZSAD methods: CLIP \citep{radford2021learning}, WinCLIP \citep{jeong2023winclip}, APRIL-GAN \citep{chen2023zero}, AnomalyCLIP \citep{zhou2023anomalyclip}, AdaCLIP \citep{cao2024adaclip} and AA-CLIP \citep{ma2025aa}. Additional details regarding the datasets and methodological specifics are provided in Appendix B.

\paragraph{Metrics} 
We adopt standard metrics for evaluation: for image-level anomaly detection, we report the Area Under the Receiver Operating Characteristic curve (AUROC), Average Precision (AP), and maximum F1 (F1); for pixel-level anomaly segmentation, we report AUROC, the Area Under the Per-Region Overlap curve (AUPRO), and F1.
\begin{figure*}[tp!]
\centering
\includegraphics[width=0.95\linewidth]{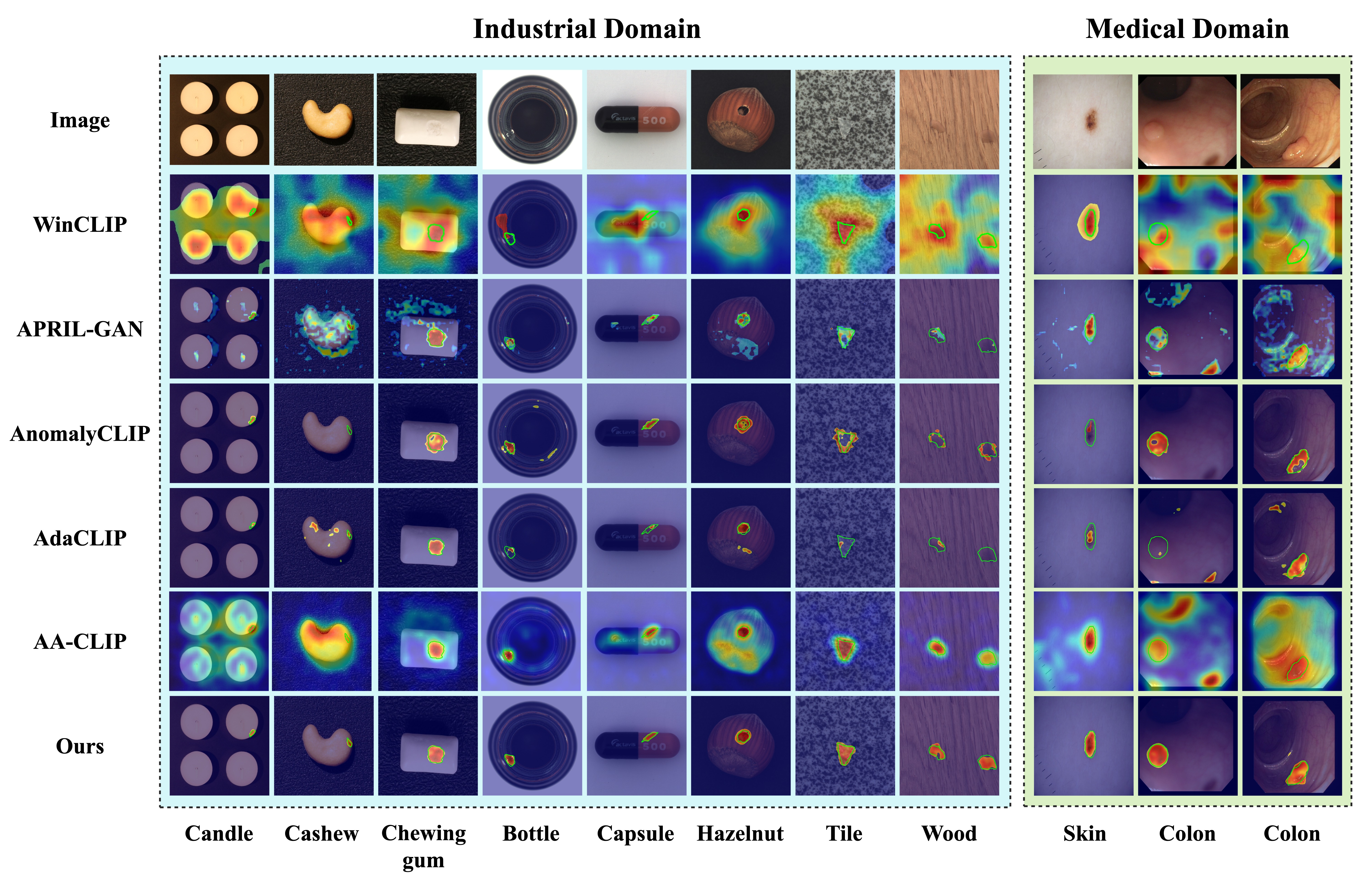}  
\caption{Visualization of anomaly maps of different ZSAD methods. Our proposed ViP$^{2}$-CLIP achieves the sharpest segmentations, capturing fine-grained defects in both industrial and medical datasets.}
\label{fig:figure4}
\end{figure*}
\paragraph{Implementation Details} 
We use the publicly released CLIP (ViT-L/14@336px) as the frozen backbone. The learnable prompts contain 10 tokens, 3 of which interact explicitly with the global visual embedding. In UTPA, we retain the top 50 anomalous patches at each alignment layer as image-level descriptors and, following prior work \citep{zhou2023anomalyclip, cao2024adaclip}, apply cross-modal alignment at layers 6, 12, 18, and 24. We fine-tune ViP$^{2}$-CLIP on MVTec AD's test split and evaluate ZSAD performance on all other datasets; for MVTec AD, we fine-tune on VisA's test set. Dataset-level metrics are averaged across all subclasses. All experiments are implemented in PyTorch 2.6.0 on a single NVIDIA L20 (48 GB). Further implementation details are provided in Appendix B.

\subsection{Main Results}
\paragraph{Zero‐Shot Anomaly Detection on Industrial Datasets}  
\cref{tab:table1} compares ViP$^{2}$-CLIP against six representative baselines on seven industrial defect benchmarks. Our model delivers competitive results and outperforms all baselines in most datasets. The vanilla CLIP struggles, as its pre-training emphasizes general object semantics rather than anomaly patterns. WinCLIP and APRIL-GAN improve detection through handcrafted prompts and local features tuning. AnomalyCLIP employs object-agnostic prompts but remains limited by static prompt design and shallow cross-modal fusion. AA-CLIP directly introduces adapter modules to enlarge the semantic gap between normal and anomalous texts, thereby enhancing detection performance. AdaCLIP refines prompts in both visual and textual space, which boosts F1 but neglects AUPRO, leading to an imbalanced ZSAD. In contrast, ViP$^{2}$-CLIP yields consistent improvements across all metrics, achieving state-of-the-art performance. Our ViP-Prompt creates fine-grained normal and abnormal prompts that align tightly with object features; paired with UTPA's unified alignment strategy, substantially improves overall ZSAD performance.
\begin{figure}[!t]
\centering
\includegraphics[width=1.0\linewidth]{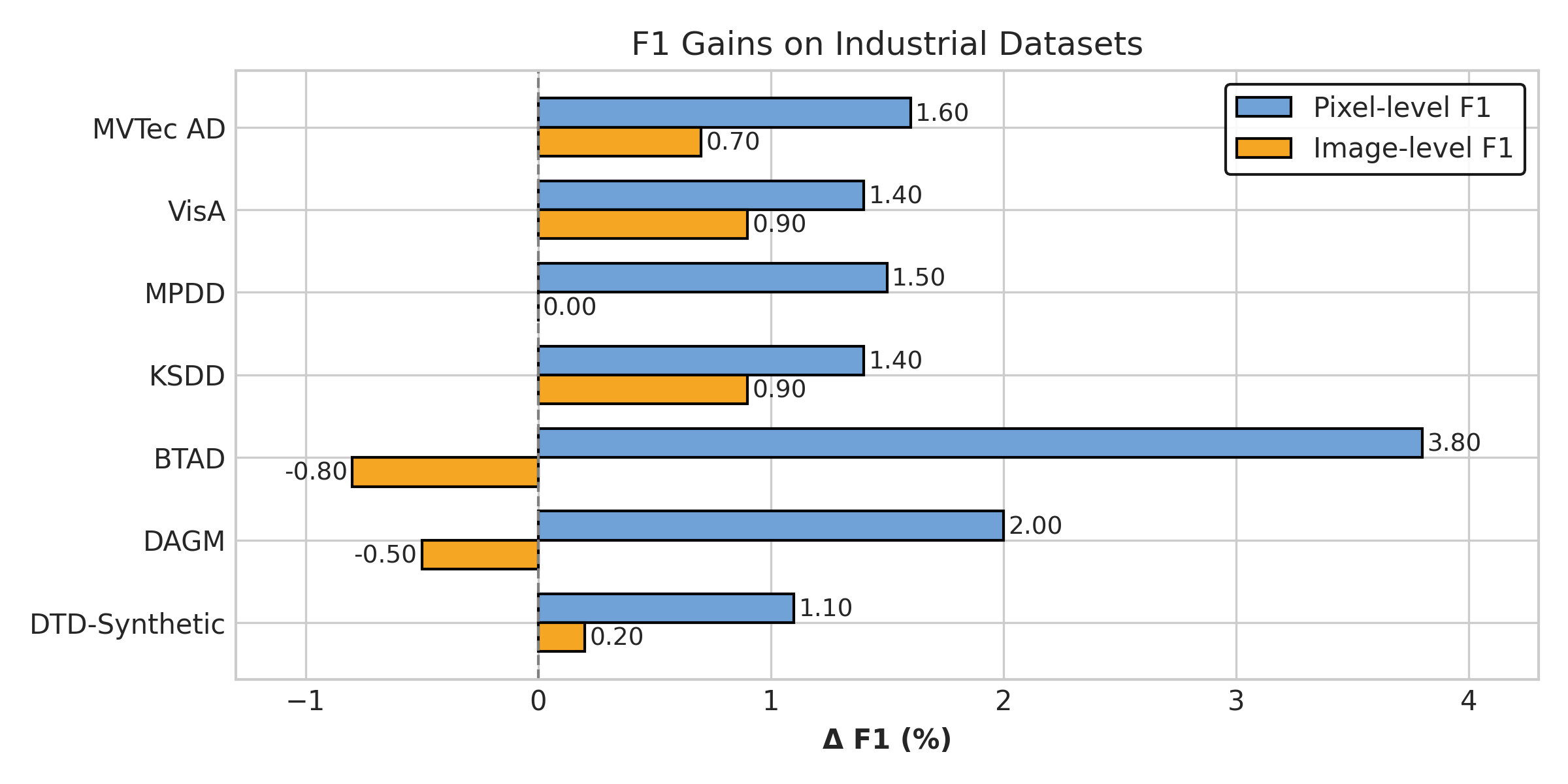}  
\caption{F1 gains of using visual-conditioned prompts compared to static learnable prompts.}
\label{fig:figure5}
\end{figure}
\cref{fig:figure4} further compares the anomaly detection results of our approach with five strong baselines, showing that ViP$^{2}$-CLIP more accurately captures defects with sharper boundaries, supporting the observed quantitative gains. We also provide failure-case analyses in Appendix C.

\paragraph{Zero‐Shot Anomaly Detection on Medical Datasets}  
To assess cross-domain transferability, we benchmark ViP$^{2}$-CLIP on eight public medical defect datasets. As shown in \cref{tab:table2}, although AA-CLIP achieves stronger pixel-level anomaly localization, its weaker image-level detection results in unbalanced ZSAD performance. In contrast, our ViP$^{2}$-CLIP consistently delivers superior results across most datasets. Qualitative examples in \cref{fig:figure4} also support this claim: our model precisely localizes melanoma lesions in dermoscopy images and colonic polyps in endoscopic frames, demonstrating robust adaptability to disparate pathological patterns.

\paragraph{Static prompts vs.\ Visual-Perception prompts}
To quantify the benefit of visual conditioning, we compare ViP$^{2}$-CLIP against ViP$^{2}$-CLIP\_re, which only uses static learnable prompts without visual cues. \cref{fig:figure5} reports image-level and pixel-level F1 gains on the seven industrial datasets. ViP$^{2}$-CLIP achieves higher pixel‐level F1 on all datasets and improves image‐level F1 on five of them. The slight drops on BTAD and DAGM are attributed to significant domain shifts from MVTec AD: BTAD reflects real-world industrial settings with complex lighting conditions, while DAGM contains synthetic textures characterized by small, repetitive anomalies. Such shifts likely distort global visual embeddings, reducing the transferability of ICA and resulting in imbalanced performance. Despite minor image-level declines on a few datasets, ViP-Prompt consistently achieves SOTA performance across diverse detection scenarios. It effectively mitigates segmentation fluctuations arising from uncertain class names in prompts under data privacy constraints (as detailed in Appendix~A), thereby ensuring more robust generalization.

\paragraph{Computational Efficiency Analysis}
\cref{tab:table_13} compares the computational efficiency with the top three baselines. AnomalyCLIP modifies self-attention weights via DPAM, while AdaCLIP introduces additional learnable tokens and a multi-scale clustering module, both of which increase computational overhead. Although AA-CLIP attains the shortest inference time through a simplified network design, its large number of trainable parameters diminishes overall efficiency. In contrast, ViP$^2$‑CLIP employs only two lightweight adapters outside the frozen CLIP backbone, enabling the fastest training (0.54 h), competitive inference speed (48.7 ms/image), and the lowest GPU memory consumption (2.21 GB), making it highly suitable for deployment in resource-constrained industrial environments.

\input{sec/table/table_22}

\subsection{Ablation Study \& Discussion}
\label{sec:ablation}
\paragraph{Module Ablation}  
We isolate the contributions of three key components: ViP-ICA, ViP-FGP, and UTPA. \cref{tab:table_3_new} shows that adding each module yields a clear, consistent gain, validating their standalone effectiveness. Specifically, ICA adaptively injects global visual context into prompt embeddings, while FGP incorporates fine‐grained local cues, enabling the prompts to capture precise irregularities. Furthermore, UTPA consistently improves both detection and localization performance upon integration, demonstrating the general effectiveness of the unified patch-level alignment strategy in guiding prompt learning toward more discriminative representations.

\input{sec/table/table_23}
\begin{figure}[!t]
\centering
\includegraphics[width=1.0\linewidth]{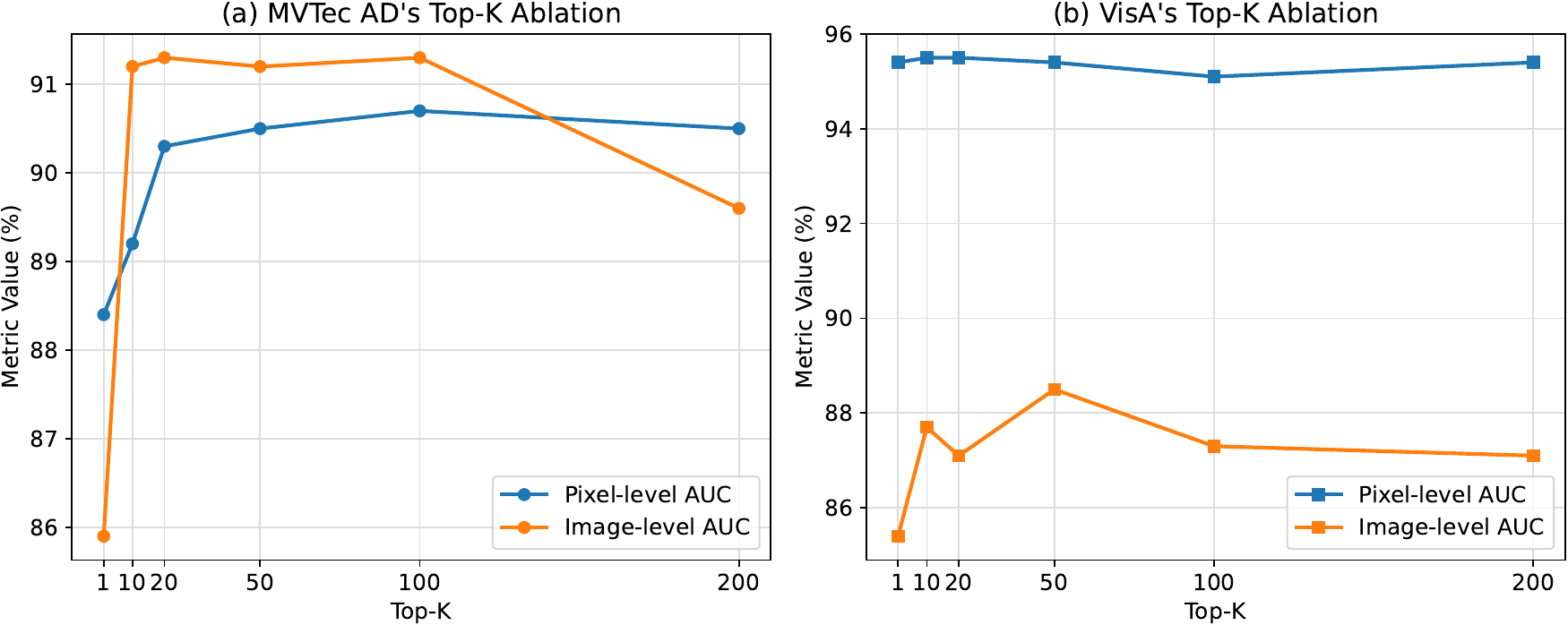}  
\caption{Top-k component ablation.}
\label{fig:figure6}
\end{figure}
\paragraph{UTPA for Resolving Optimization Conflicts}
We conduct an ablation study using a frozen CLIP backbone with learnable prompts for normal and anomalous classes to evaluate the effectiveness of UTPA in guiding model learning. The baseline (CLIP\_DUAL) adopts a dual-branch strategy that aligns prompts with both global and local visual cues. In contrast, CLIP\_UTPA simplifies the optimization by aligning solely with local features. As shown in \cref{tab:table12}, on the MVTec AD dataset, the dual-branch strategy tends to overfit image-level scores while underperforming on pixel-level detection. UTPA, by comparison, achieves more stable and superior results. As for VisA, it outperforms the baseline across both detection metrics. Several results in \cref{tab:table_3_new} also support this observation: traditional dual-branch approaches often suffer from imbalanced performance due to conflicting objectives, whereas UTPA alleviates this issue through consistent text-patch alignment. This facilitates more robust model learning and provides a novel alignment paradigm for more accurate ZSAD research.

\paragraph{Top‐K Ablation} 
To assess how the image-level representation is affected by the number of anomalous patches, we vary $K\in\{1,10,20,50,100,200\}$ and examine the AUROC performance in \cref{fig:figure6}. At $K=1$, the model is overly sensitive to local noise, yielding a low AUROC. As $K$ increases, performance steadily improves, demonstrating that aggregating multiple high-scoring anomalous patches enhances robustness. On MVTec AD, performance plateaus when $K$ is at 50 or 100; As for VisA, the optimum occurs at $K=50$. Beyond these points, additional patches introduce mostly normal regions, diluting the anomaly signals and reducing accuracy. Balanced across all metrics, we select the top 50 most anomalous patches as image-level descriptors, which effectively capture the anomaly characteristics in most detection scenarios.

\begin{figure}[!t]
\centering
\includegraphics[width=1.0\linewidth]{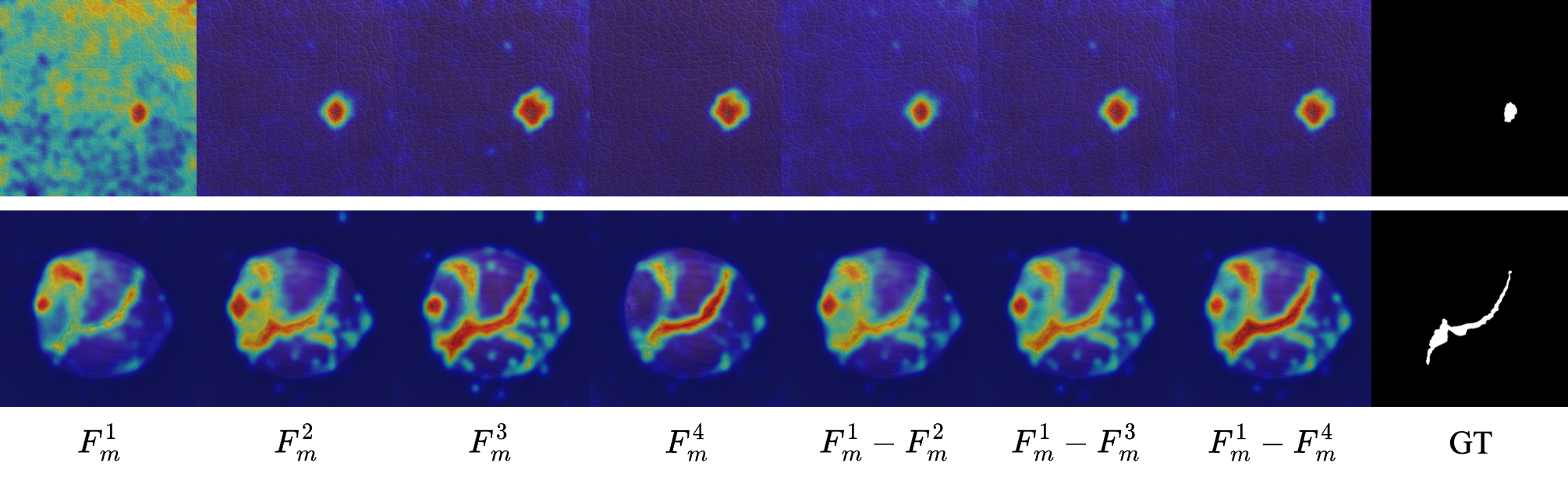}  
\caption{Visualization of anomaly maps from different layers.}
\label{fig:figure11}
\end{figure}

\paragraph{Visualization at Different layers}
To examine the impact of model depth on anomaly representation, \cref{fig:figure11} presents anomaly maps derived from $F_m^1$ to $F_m^4$ layers. For complex object categories such as hazelnut, deeper layers yield more accurate anomaly localization. In contrast, for simpler texture-based objects like leather, intermediate layers offer sufficient discriminative power. Aggregating features across multiple depths provides complementary contextual cues, allowing the model to better adapt to diverse anomaly types, and thereby achieve more robust anomaly detection.

\input{sec/table/table_21}

%% file: sec/table/table_22.tex
\begin{table}[]
\resizebox{\linewidth}{!}{
\begin{tabular}{lcccc}
\hline
\noalign{\hrule height 0.5mm}
Model       & Training Time (h) & Inference Time (ms) & GPU Cost (GB) & Trainable Parameters (M) \\ \hline
AnomalyCLIP & 1.02              & 90.72 ± 0.16        & 2.75                   & 5.56                      \\
AdaCLIP     & 2.24              & 134.20 ± 0.39       & 3.17                   & 10.67                      \\
AA-CLIP      & 0.93              & \textcolor{red}{19.17 ± 0.23}        & 2.87                   & 12.58  \\
ViP$^2$-CLIP & \textcolor{red}{0.54} & 48.69 ± 0.19 & \textcolor{red}{2.21} & \textcolor{red}{4.15} 
                    \\ \hline
\noalign{\hrule height 0.5mm}
\end{tabular}
}
\caption{Comparison of computational efficiency on VisA.}
\label{tab:table_13}
\end{table}

%% file: sec/table/table_23.tex
\begin{table}[]
\centering
\resizebox{\linewidth}{!}{
\begin{tabular}{ccccccccccc}
\hline
\noalign{\hrule height 0.5mm}
\multirow{2}{*}{{UTPA}} & \multicolumn{2}{c}{ViP-Prompt} & \multicolumn{4}{c}{MVTec AD} & \multicolumn{4}{c}{VisA} \\
 & {ICA} & {FGP} & \multicolumn{2}{c}{Pixel-level} & \multicolumn{2}{c}{Image-level} & \multicolumn{2}{c}{Pixel-level} & \multicolumn{2}{c}{Image-level} \\
\hline
                      &                      &                       & (37.8, 11.5, 7.0) &        & (74.1, 87.6, 87.2) &        & (43.5, 14.6, 2.8) &        & (60.2, 66.3, 74.5) &        \\
                      & \checkmark           &                       & (89.5, 79.9, 38.6) &        & (71.8, 86.2, 88.5) &        & (95.2, 90.9, 31.6) &        & (79.7, 82.8, 79.5) &        \\
                      &                      & \checkmark           & (90.0, 85.5, 42.7) &        & (66.1, 82.6, 86.6) &        & (95.3, 91.5, 33.4) &        & (60.1, 66.4, 73.2) &        \\
                      & \checkmark           & \checkmark           & (87.7, 83.8, 42.4) &        & (88.9, 94.9, 91.6) &        & (95.2, 91.6, \textcolor{red}{35.5}) &        & (83.0, 85.9, 81.4) &        \\
\checkmark           & \checkmark           &                       & (89.8, 84.4, 40.2) &        & (89.2, 95.0, 91.4) &        & (94.9, 91.4, 31.3) &        & (86.3, 89.1, 83.8) &        \\
\checkmark           &                      & \checkmark           & (88.8, 82.1, 38.7) &        & (85.2, 93.3, 89.0) &        & (95.3, 91.9, 31.4) &        & (87.6, 89.4, 83.8) &        \\
\checkmark           & \checkmark           & \checkmark           & (\textcolor{red}{90.5}, \textcolor{red}{87.1}, \textcolor{red}{43.1}) &        & (\textcolor{red}{91.2}, \textcolor{red}{96.0}, \textcolor{red}{92.0}) &        & (\textcolor{red}{95.4}, \textcolor{red}{92.2}, 33.6) &        & (\textcolor{red}{88.5}, \textcolor{red}{90.4}, \textcolor{red}{84.8}) &        \\
\hline
\noalign{\hrule height 0.5mm}
\end{tabular}
}
\caption{Module ablation.}
\label{tab:table_3_new}
\end{table}

%% file: sec/table/table_21.tex
\begin{table}[]
\resizebox{\linewidth}{!}{
\begin{tabular}{lcccc}
\hline
\noalign{\hrule height 0.5mm}
\multirow{2}{*}{Module} & \multicolumn{2}{c}{MVTec AD}                             & \multicolumn{2}{c}{VisA}                             \\
                        & Pixel-level               & Image-level              & Pixel-level               & Image-level              \\ \hline
CLIP\_DUAL           & (69.6, 24.6, 12.0)         & \textcolor{red}{(88.6, 94.9, 91.3)}        & (93.3, 82.4, 24.2)         & (79.3, 82.4, 79.0)        \\
CLIP\_UTPA              & \textcolor{red}{(89.6, 83.7, 37.4)}         & (85.0, 93.3, 88.8)        & \textcolor{red}{(94.4, 88.4, 26.2)}         & \textcolor{red}{(83.8, 86.7, 81.4)}        \\ \hline
\noalign{\hrule height 0.5mm}
\end{tabular}
}
\caption{Ablation on the effectiveness of UTPA.}
\label{tab:table12}
\end{table}

%% file: sec/6_conclusion.tex
\section{Conclusion}

We propose ViP$^2$‑CLIP, a universal ZSAD framework that detects anomalies in unseen categories without any target domain training samples. At its core, ViP‑Prompt adaptively integrates global and local visual cues into text prompts, eliminating reliance on manual templates or class-name priors, achieving superior generalization and robustness in cross-modal alignment. In addition, the UTPA optimizes a unified text-patch alignment strategy across scales, sharpening both detection and localization performance. Extensive experiments on 15 mainstream datasets demonstrate that ViP$^2$‑CLIP consistently outperforms existing methods across various detection scenarios, especially when category labels are ambiguous or privacy-constrained, offering a more accurate and scalable solution for ZSAD.

\paragraph{Limitations} 
Although UTPA achieves high-precision ZSAD performance through a simple yet effective alignment strategy, its current implementation only selects the better-performing Top-K from a small set of discrete candidates, further work on more advanced feature aggregation strategies is needed to explore.

%% file: sec/7_Appeix.tex
\appendix
\section{Motivation Statement}
\label{sec:appendixa}

\subsection{Robustness Analysis of Class-Name in Prompts}
\begin{figure}[h!]
\centering
\includegraphics[width=1.0\linewidth]{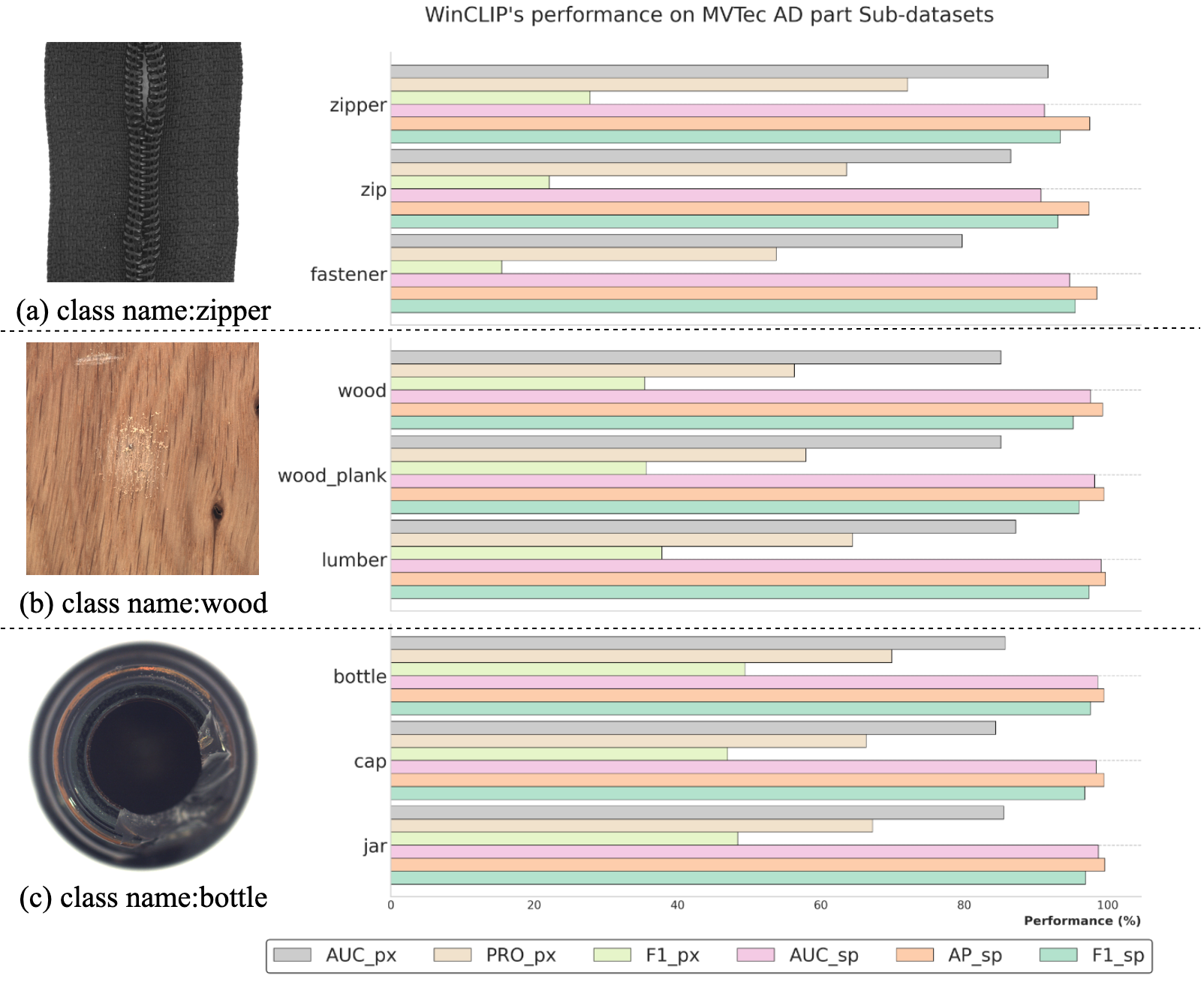} 
\caption{WinCLIP's performance of different Class Names in Prompts.}
\label{fig:figure7}
\end{figure}

Methods like CLIP-AD \citep{chen2024clip}, AdaCLIP \citep{cao2024adaclip}, and KANOCLIP \citep{li2025kanoclip} enhance alignment by embedding class names into prompts, but this introduces a critical vulnerability: CLIP's segmentation quality is highly sensitive to label phrasing. Specifically, we conducted synonym-swap experiments on the training-free WinCLIP model to evaluate how CLIP's inherent segmentation capability is influenced by the class-name variations. As shown in \cref{fig:figure7}, by replacing class names with semantically equivalent variants (e.g., `zipper' to `zip' or `fastener'), we observed a striking inconsistency: pixel-level segmentation metrics can vary by up to 10 percentage points, while image-level classification metrics remain stable. We attribute this instability to CLIP's classification-oriented pre-training, which enforces robust global alignment but yields fragile local alignment. 

\begin{figure}[h!]
\centering
\includegraphics[width=1.0\linewidth]{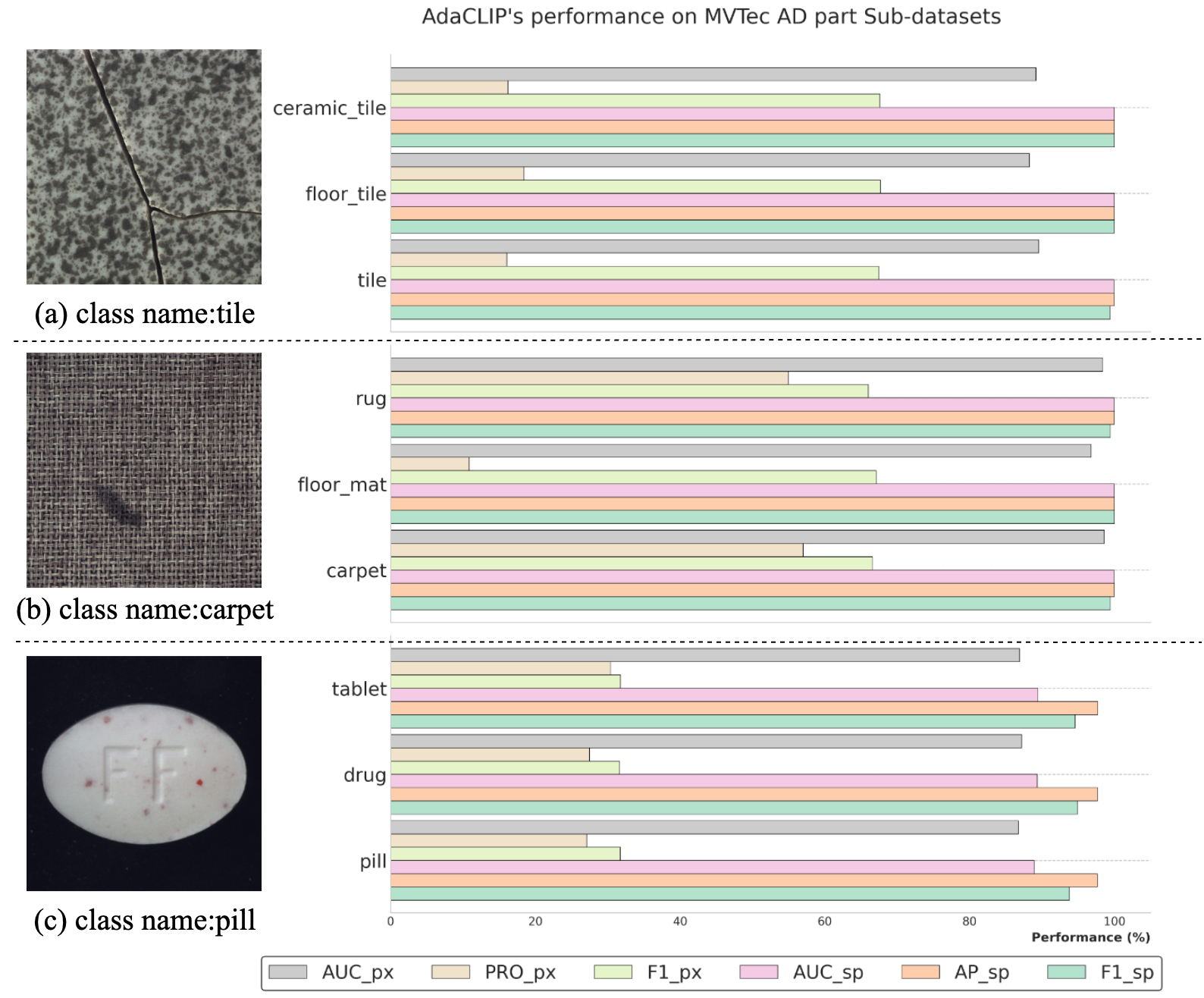}  
\caption{AdaCLIP's performance of different Class Names in Prompts.}
\label{fig:figure8}
\end{figure}

To further evaluate whether training-based modality alignment can alleviate this sensitivity, we repeated the synonym-swap experiments on recent work, AdaCLIP. As shown in \cref{fig:figure8}, although AdaCLIP's pixel-level F1 stays stable, its AUROC and PRO fluctuate with label swaps; in particular, changing `carpet' to `floor mat' triggers a 47\% swing in PRO. This dramatic fluctuation shows that, even with fine-tuning on auxiliary data, the limited scale of anomaly-specific training cannot effectively mitigate CLIP's segmentation sensitivity to class-name perturbations. These findings underscore the necessity of label-agnostic prompt designs for truly robust ZSAD.

\section{Experimental Details}
\label{sec:appendixb}
\input{sec/table/table_5}
\subsection{Implementation Details}
We use the publicly released CLIP (ViT‑L/14@336px) model as the frozen feature extractor. In all experiments, the learnable prompt length is set to 10, of which 3 tokens fuse with the global visual feature. In UTPA, we set $Top‑k=50$ and perform cross-modal alignment at layers $\{6,12,18,24\}$. During training, input images are resized to $518\times518$, we optimize with Adam at a learning rate of $1\times10^{-3}$ for 10 epochs with a batch size of 8. All experiments are implemented in PyTorch 2.6.0 on a single NVIDIA L20 (48 GB).

We conduct comprehensive evaluations across 15 publicly available benchmarks spanning industrial and medical domains. Specifically, we fine‑tune on MVTec AD's test set and perform zero‑shot evaluation on all other datasets; for MVTec AD, we fine‑tune on VisA's test split. We adopt standard metrics for evaluation: Area Under the ROC Curve (AUC), Average Precision (AP), and F1-score (F1) for image-level anomaly detection; AUC, Per-Region Overlap (PRO), and F1 for pixel-level segmentation.

\subsection{Baselines}
To validate the superiority of ViP$^2$‑CLIP, we compare against the following state‑of‑the‑art methods:

\begin{itemize}
  \item \textbf{CLIP} \citep{radford2021learning}: We adapt CLIP for anomaly detection by using two prompt templates—`a photo of a normal [cls]' and `a photo of an abnormal [cls]', where \texttt{[cls]} is the category class-name. We compute image‑level anomaly scores via cosine similarity between global visual features and text embeddings. For pixel‑level segmentation, we extend this scoring to the last‑layer patch embeddings.
  \item \textbf{WinCLIP} \citep{jeong2023winclip}: WinCLIP is the first CLIP‑based ZSAD method, which uses multiple manually designed prompts and a multi‑window feature extraction strategy. All parameters are kept the same as in their paper.
  \item \textbf{APRIL‑GAN}\citep{chen2023zero}: APRIL‑GAN extends WinCLIP by incorporating learnable linear projection layers to enrich local visual feature representations. All experiments are conducted using the official weights and adhere strictly to the original implementation protocols.
  \item \textbf{AnomalyCLIP} \citep{zhou2023anomalyclip}: AnomalyCLIP introduces object‑agnostic learnable prompts and a DPAM mechanism for stronger local features' modelling. All experiments are conducted using publicly released official model weights.
  \item \textbf{AdaCLIP} \citep{cao2024adaclip}: AdaCLIP incorporates hybrid prompts into both text and image encoders with an HSF module for improved cross‑modal fusion. As AdaCLIP adopts an evaluation protocol distinct from ours, it trains on MVTec AD and CVC-ColonDB and performs zero-shot testing on the remaining datasets. To ensure fair and consistent comparison, we reimplement AdaCLIP under our unified evaluation protocol. All parameters are kept the same as in their paper.
  \item \textbf{AA-CLIP} \citep{ma2025aa}: AA-CLIP enhances anomaly detection by adopting a two-stage training strategy that enlarges the semantic gap between normal and anomalous texts through joint optimization of the text and image encoders. We reimplement it under our unified evaluation protocol, keeping all parameters consistent with the original paper.
\end{itemize}

\input{sec/table/table_6}

\subsection{Datasets}
We evaluate on 7 industrial datasets (MVTec AD, VisA, MPDD, BTAD, KSDD, DAGM, DTD‑Synthetic) and 8 medical datasets (HeadCT, BrainMRI, Br35H, ISIC, CVC‑ClinicDB, CVC‑ColonDB, Endo, Kvasir), for a total of 15 benchmarks. All methods are trained and tested solely on each dataset's test split. The dataset statistics are given in \cref{tab:table5}. We apply OpenCLIP's default normalization to all images and resize them to $(518,518)$ to obtain suitably scaled feature maps.

\section{Additional Experiments and Analysis}
\label{sec:appendixc}
\subsection{Comparison with Full-shot Methods}
In this section, we compare ViP$^2$‑CLIP against two state‑of‑the‑art full-shot methods, PatchCore \citep{roth2022towards} and RD4AD \citep{deng2022anomaly}, on five industrial datasets for which training samples are available. Results are presented in \cref{tab:table6}. Despite not using any target domain training samples, ViP$^2$‑CLIP achieves detection and segmentation performance on par with or superior to these fully supervised approaches, with particularly strong gains on the BTAD and DAGM datasets. This demonstrates that ViP$^2$‑CLIP's visual‑perception prompt mechanism adaptively generates fine‑grained normal and anomalous text descriptions that fully unleash CLIP's ZSAD performance, breaking the dependency of traditional methods on in‑domain training data and exhibits excellent generalization.
\input{sec/table/table_8}

\subsection{Targeted Fine‑tuning on Medical Data}
Although ViP$^2$‑CLIP achieves strong results on industrial benchmarks, its performance on medical images is relatively weaker, partly due to the domain gap between industrial data and medical target domains. To investigate whether fine‑tuning on medical auxiliary data can bridge this gap, we design a cross‑domain fine-tuning protocol: we train on MVTec AD and CVC-ColonDB test split and test on the remaining medical datasets; for CVC-ColonDB, we use VisA and CVC-ClinicDB test sets as auxiliary data. We compare against AnomalyCLIP under the same training strategy. As shown in \cref{tab:table8}, introducing medical auxiliary data significantly boosts detection performance, particularly on CVC-ClinicDB and Endo datasets, highlighting the critical role of auxiliary data in cross‑domain generalization. Overall, ViP$^2$‑CLIP outperforms AnomalyCLIP under medical fine‑tuning, demonstrating superior cross‑domain anomaly detection capability.

\begin{figure}[tp!]
\centering
\includegraphics[width=1.0\linewidth]{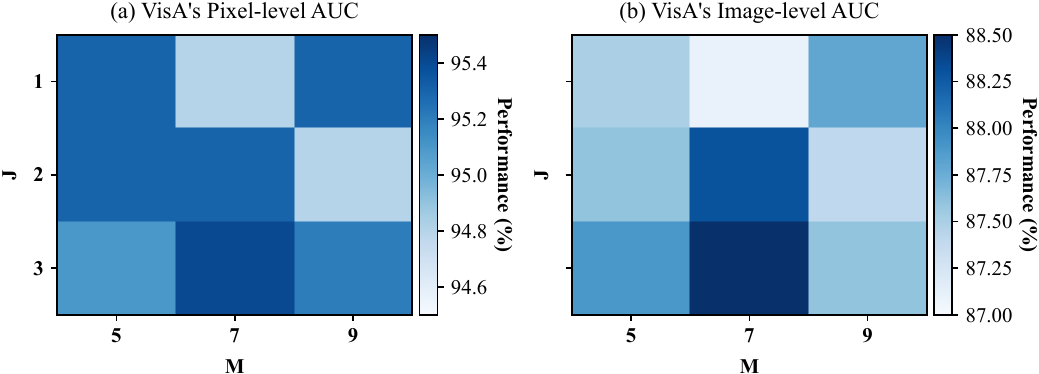}  
\caption{Prompt Length ablation.}
\label{fig:figure9}
\end{figure}

\input{sec/table/table_4}
\subsection{State‐Adjective Ablation}  
To verify our prompt's robustness, we replace the original `good/damaged' adjectives with several alternative pairs of similar meaning and retrain the model on MVTec-AD and VisA. As \cref{tab:table4} shows, detection performance remains consistent across all variants, demonstrating insensitivity to specific adjective choices. This resilience stems from our learnable prompt tokens, which autonomously capture normal and anomalous semantics, obviating the need for handcrafted text templates.

\subsection{Prompt Length Ablation}  
Next, based on VisA dataset, we vary the total prompt length to assess its effect on ZSAD performance, focusing on both the number of static learnable tokens $M$ and dynamic tokens $J$, which fuse with global visual context. \cref{fig:figure9} shows that increasing length does not always improve performance: too many learnable tokens introduce redundancy and risk overfitting, harming precision and generalization capacity. To balance between semantic expressiveness and model robustness, we set the prompt length to 10 by default, comprising 3 dynamic tokens infused with global visual semantics and 7 static tokens modelling generic normal and anomalous patterns.

\subsection{Failure Cases}  
ViP$^2$‑CLIP performs well in clean and controlled environments with localized anomalies, but certain failure cases arise in more complex scenarios:

\paragraph{Contextual Anomalies}
The model struggles to detect contextual anomalies—such as abnormal spatial arrangements or misaligned components—that lack explicit visual distortions. These cases remain challenging without prior knowledge of normal structural configurations. This highlights a promising direction: integrating object-level positional priors or spatial layout descriptions into prompt design to enhance contextual reasoning.

\paragraph{Highly Textured Backgrounds}
Our method assumes that the target object is centrally located within a clean background, allowing the image encoder to extract a reliable global visual embedding for prompt generation. However, in cluttered scenes or when multiple objects are present, the quality of the extracted global embedding deteriorates. The absence of class name guidance further impairs localization, reducing detection accuracy. 

\paragraph{Uneven Anomaly Distributions}
In UTPA, a fixed Top-K patch aggregation strategy is used to represent global anomaly features. However, when anomalies are unevenly distributed across the image, the choice of K affects performance. While our Top-50 selection achieves SOTA results under the detection settings of most datasets, future work should explore adaptive aggregation mechanisms to better accommodate varying anomaly distributions.

\section{Additional Quantitative and Qualitative Results}
\label{sec:appendixd}
\subsection{Fine‑grained Subset Performance}
\cref{tab:table9}--\cref{tab:table20} report detailed ZSAD performance at the subset level on MVTec AD and VisA datasets, illustrating ViP$^2$‑CLIP's ability to handle fine‑grained variations within each category.

\subsection{Visualization}
\begin{figure*}[htbp]
\centering
\includegraphics[width=1.0\linewidth]{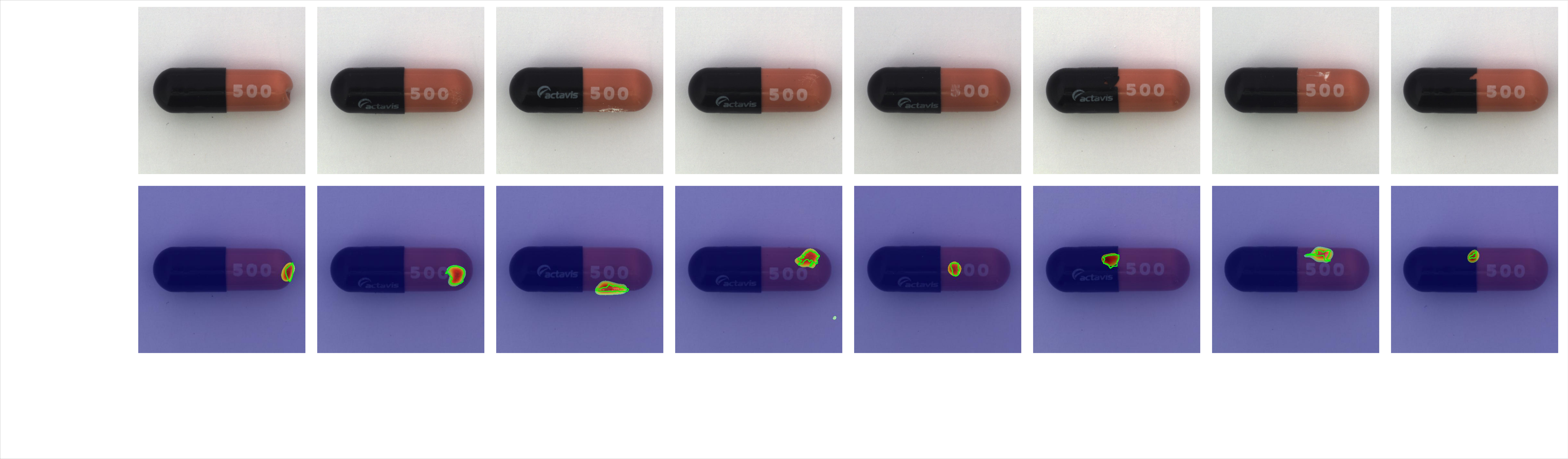} 
\caption{Anomaly score maps for the data subset, {capsule}, in {MVTec AD}. The first row represents the input, the second row presents the segmentation results from ViP$^2$-CLIP.}
\label{fig:figure10}
\end{figure*}

\begin{figure*}[htbp]
\centering
\includegraphics[width=1.0\linewidth]{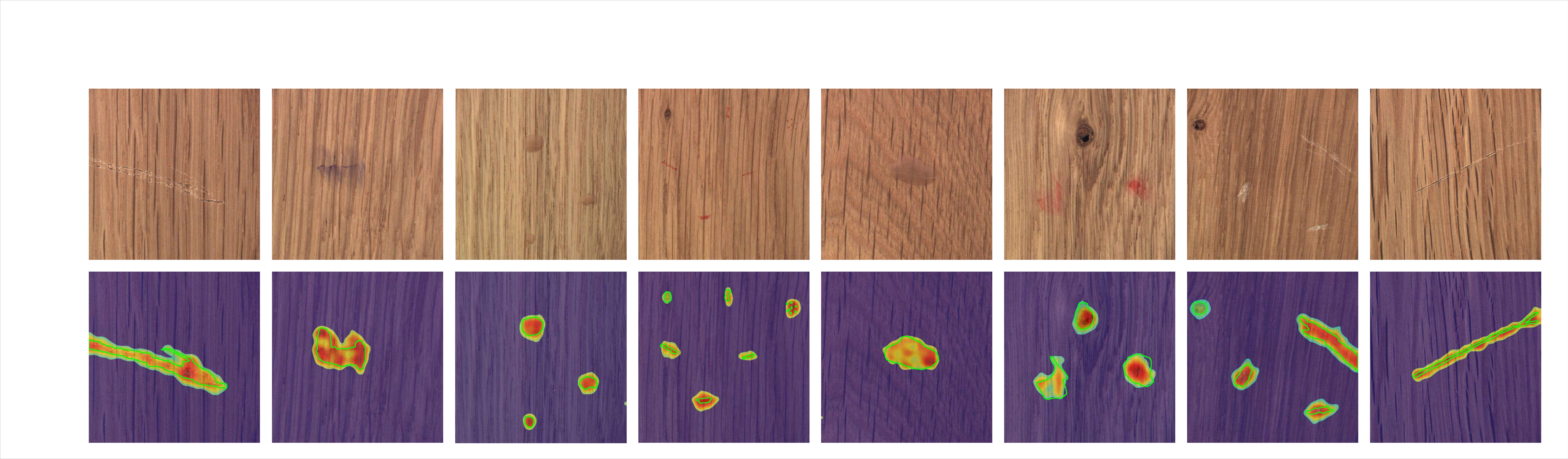} 
\caption{Anomaly score maps for the data subset, {wood}, in {MVTec AD}. The first row represents the input, the second row presents the segmentation results from ViP$^2$‑CLIP.}
\label{fig:figure12}
\end{figure*}

\begin{figure*}[htbp]
\centering
\includegraphics[width=1.0\linewidth]{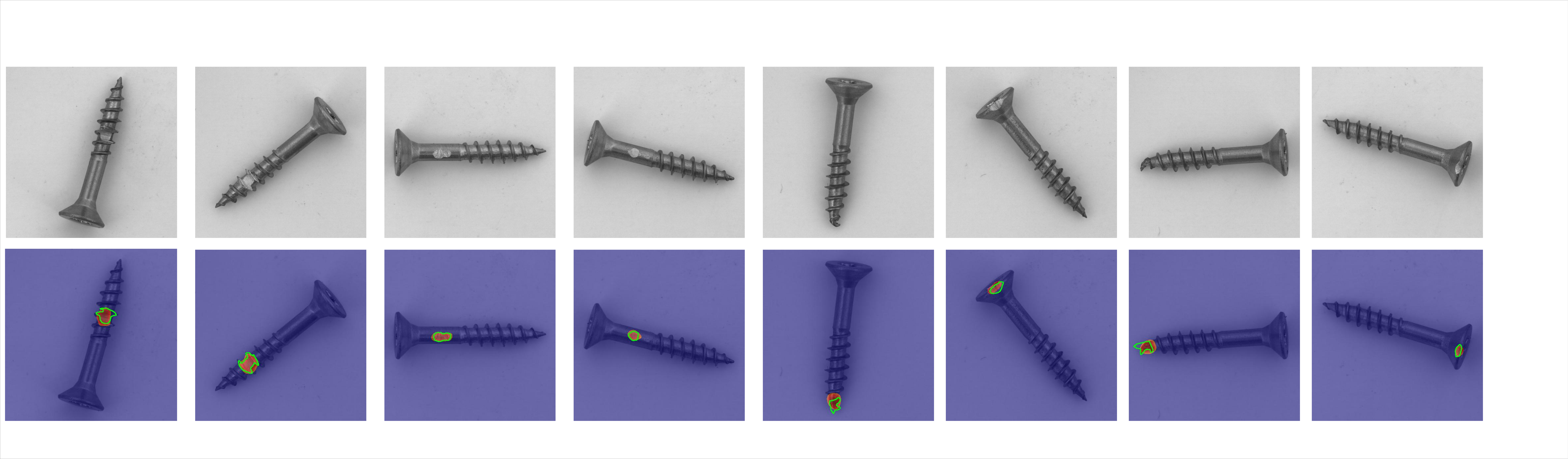} 
\caption{Anomaly score maps for the data subset, {screw}, in {MVTec AD}. The first row represents the input, the second row presents the segmentation results from ViP$^2$‑CLIP.}
\label{fig:figure13}
\end{figure*}

\begin{figure*}[htbp]
\centering
\includegraphics[width=1.0\linewidth]{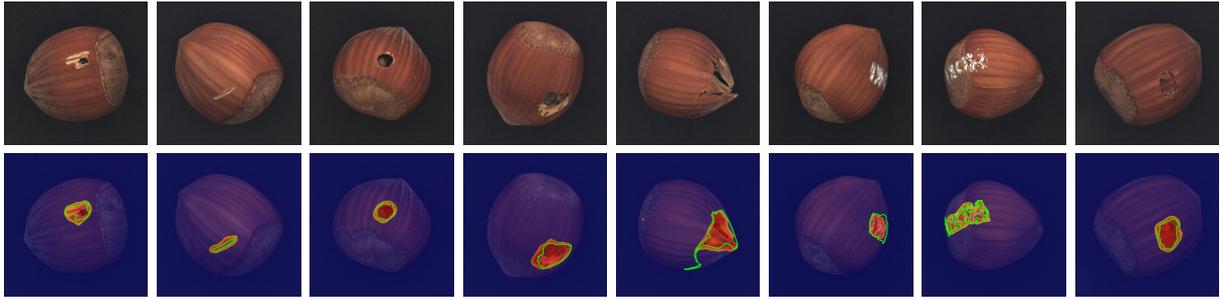} 
\caption{Anomaly score maps for the data subset, {hazelnut}, in {MVTec AD}. The first row represents the input, the second row presents the segmentation results from ViP$^2$‑CLIP.}
\label{fig:figure14}
\end{figure*}

We visualize pixel‑wise anomaly maps in \cref{fig:figure10}--\cref{fig:figure23}, demonstrating ViP$^2$‑CLIP’s strong segmentation capability and cross‑domain generalization. Industrial examples include capsule, grid, leather, screw and hazelnut from MVTec AD; candle, cashew, macaroni, pcb and pipe fryum from VisA; brackets and metal plates from MPDD; and unknown products from BTAD. Medical examples cover melanoma detection in ISIC, and colorectal polyp detection in CVC‑ColonDB and CVC‑ClinicDB.


\input{sec/table/table_9}
\input{sec/table/table_10}
\input{sec/table/table_11}
\input{sec/table/table_12}
\input{sec/table/table_13}
\input{sec/table/table_14}
\input{sec/table/table_15}
\input{sec/table/table_16}
\input{sec/table/table_17}
\input{sec/table/table_18}
\input{sec/table/table_19}
\input{sec/table/table_20}
\clearpage


\begin{figure*}[htbp]
\centering
\includegraphics[width=1.0\linewidth]{sec/pic_/visa-cashew.pdf} 
\caption{Anomaly score maps for the data subset, {cashew}, in {VisA}. The first row represents the input, the second row presents the segmentation results from ViP$^2$‑CLIP.}
\label{fig:figure16}
\end{figure*}

\begin{figure*}[htbp]
\centering
\includegraphics[width=1.0\linewidth]{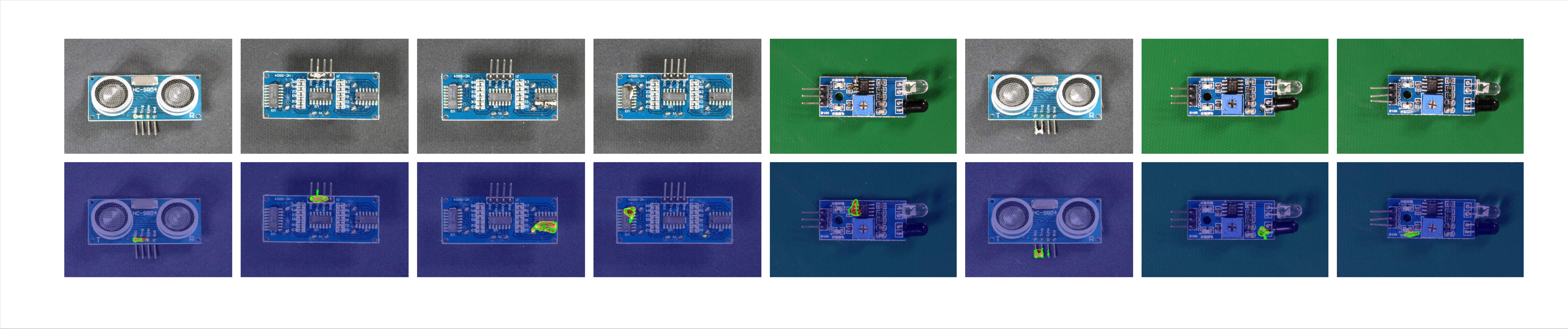} 
\caption{Anomaly score maps for the data subset, {pcb}, in {VisA}. The first row represents the input, the second row presents the segmentation results from ViP$^2$‑CLIP.}
\label{fig:figure17}
\end{figure*}

\begin{figure*}[htbp]
\centering
\includegraphics[width=1.0\linewidth]{sec/pic_/visa-pipe.pdf} 
\caption{Anomaly score maps for the data subset, {pipe fryum}, in {VisA}. The first row represents the input, the second row presents the segmentation results from ViP$^2$‑CLIP.}
\label{fig:figure18}
\end{figure*}

\begin{figure*}[htbp]
\centering
\includegraphics[width=1.0\linewidth]{sec/pic_/mpdd.pdf} 
\caption{Anomaly score maps for the data subset in {MPDD}. The first row represents the input, the second row presents the segmentation results from ViP$^2$‑CLIP.}
\label{fig:figure19}
\end{figure*}

\begin{figure*}[htbp]
\centering
\includegraphics[width=1.0\linewidth]{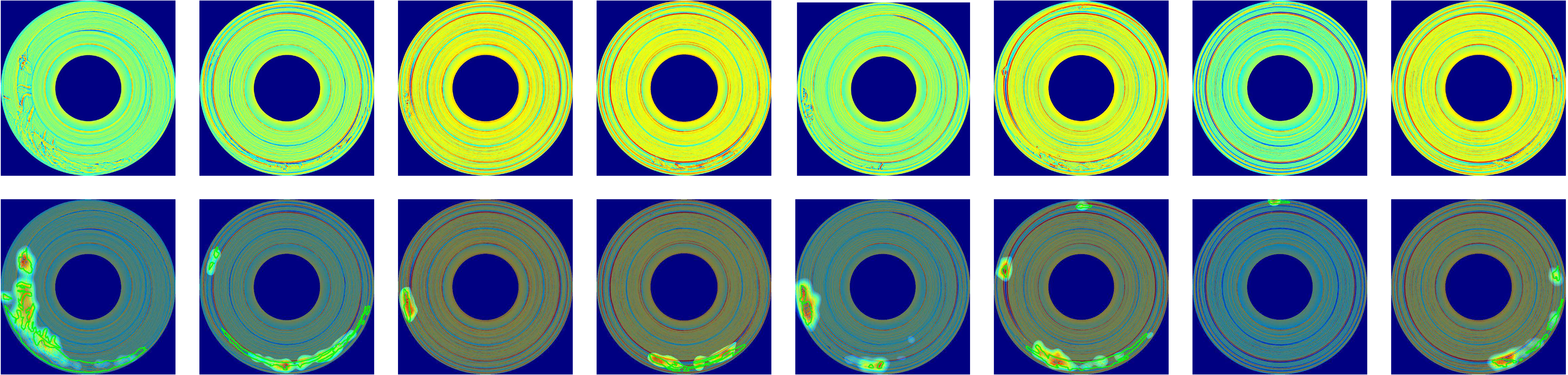} 
\caption{Anomaly score maps for the data subset in {BTAD}. The first row represents the input, the second row presents the segmentation results from ViP$^2$‑CLIP.}
\label{fig:figure20}
\end{figure*}

\begin{figure*}[htbp]
\centering
\includegraphics[width=1.0\linewidth]{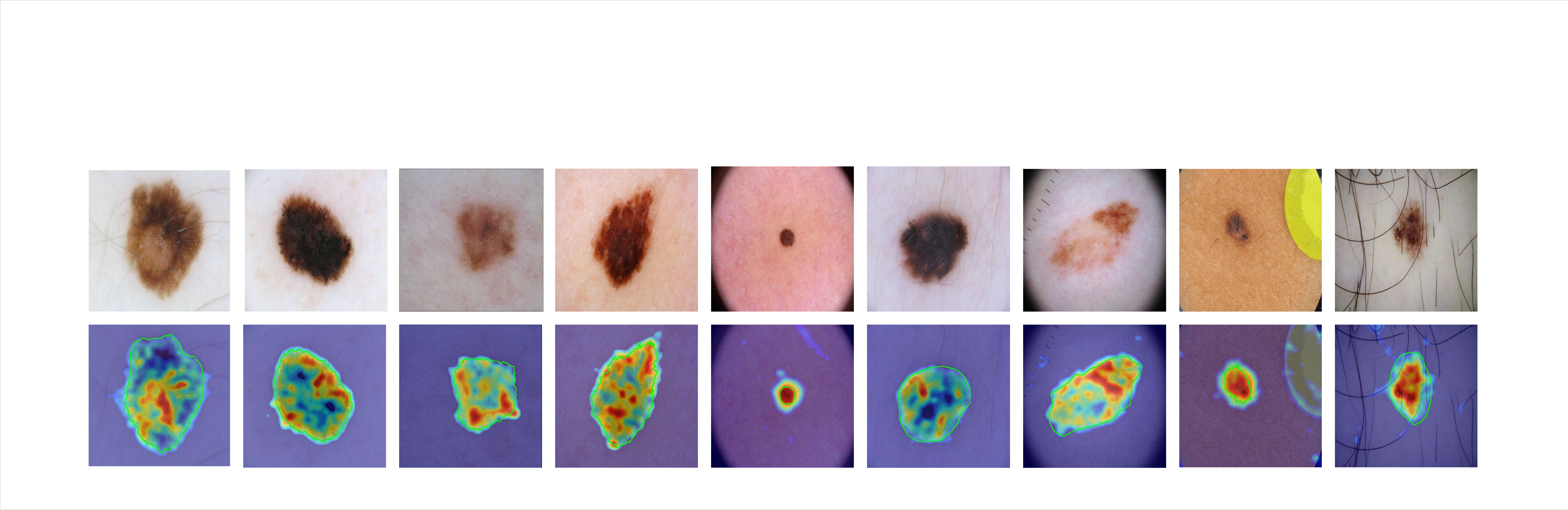} 
\caption{Anomaly score maps for the data subset skin. The first row represents the input, the second row presents the segmentation results from ViP$^2$‑CLIP.}
\label{fig:figure21}
\end{figure*}

\begin{figure*}[htbp]
\centering
\includegraphics[width=1.0\linewidth]{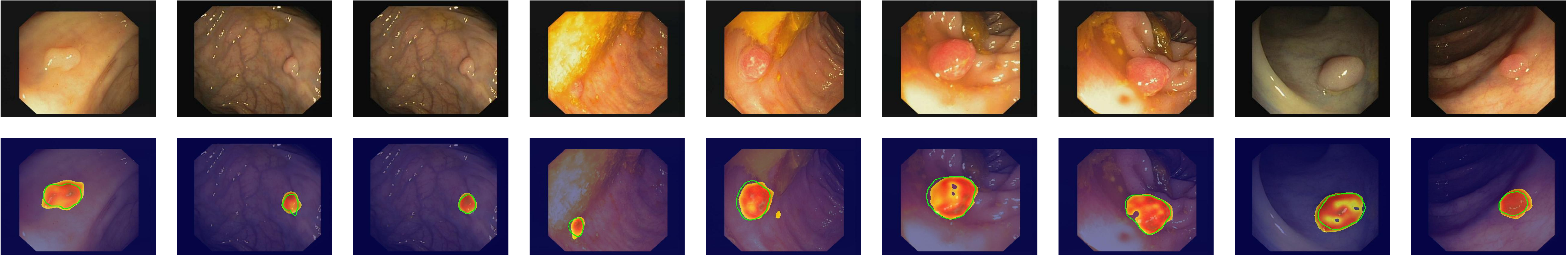} 
\caption{Anomaly score maps for the data subset colon. The first row represents the input, the second row presents the segmentation results from ViP$^2$‑CLIP.}
\label{fig:figure22}
\end{figure*}

\begin{figure*}[htbp]
\centering
\includegraphics[width=1.0\linewidth]{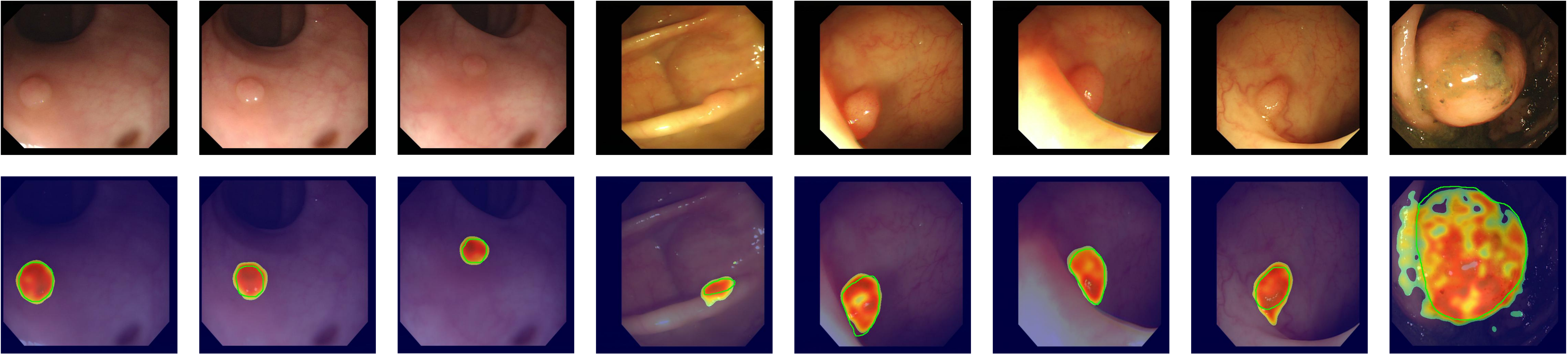} 
\caption{Anomaly score maps for the data subset colon. The first row represents the input, the second row presents the segmentation results from ViP$^2$‑CLIP.}
\label{fig:figure23}
\end{figure*}

\clearpage

%% file: sec/table/table_5.tex
\begin{table*}[]
\resizebox{\linewidth}{!}{
\begin{tabular}{cllccc}
\hline
\noalign{\hrule height 0.5mm}
Domain                        & Dataset         & Category            & Modalities       & $|\mathcal{C}|$ & Normal and anomalous samples \\ \hline
\multirow{7}{*}{Industrial} & MVTec AD        & Obj \& texture     & Photography     & 15              & (467, 1258)                  \\ \cline{2-6}
                            & VisA            & \multirow{4}{*}{Obj} & Photography     & 12              & (962, 1200)                  \\
                            & MPDD            &                   & Photography     & 6               & (176, 282)                   \\
                            & BTAD            &                   & Photography     & 3               & (451, 290)                   \\
                            & KSDD            &                   & Photography     & 1               & (286, 54)                    \\ \cline{2-6}
                            & DAGM            & \multirow{2}{*}{Texture}           & Photography     & 10              & (6996, 1054)                 \\
                            & DTD-Synthetic   &            & Photography     & 12              & (357, 947)                   \\ \hline
\multirow{8}{*}{Medical}       & ISIC            & Skin              & Photography     & 1               & (0, 379)                     \\ \cline{2-6}
& CVC-ClinicDB    & \multirow{4}{*}{Colon} & Endoscopy       & 1               & (0, 612)                     \\
                            & CVC-ColonDB     &                   & Endoscopy       & 1               & (0, 380)                     \\
                            & Kvasir          &                   & Endoscopy       & 1               & (0, 1000)                    \\
                            & Endo            &                   & Endoscopy       & 1               & (0, 200)                     \\ \cline{2-6}
                            & HeadCT          & \multirow{3}{*}{Brain} & Radiology (CT)  & 1               & (100, 100)                   \\
                            & BrainMRI        &                   & Radiology (MRI) & 1               & (98, 155)                    \\
                            & Br35H           &                   & Radiology (MRI) & 1               & (1500, 1500)                 \\ 
\noalign{\hrule height 0.5mm}
\end{tabular}
}
\caption{Statistics of the datasets.}
\label{tab:table5}
\end{table*}

%% file: sec/table/table_6.tex
\begin{table*}[ht]
\resizebox{\linewidth}{!}{
\begin{tabular}{cccccc}
\hline
\noalign{\hrule height 0.5mm}
Task                                        & Category             & Datasets & PatchCore$^{*}$   & RD4AD$^{*}$       & ViP$^{2}$-CLIP         \\ \hline
\multirow{5}{*}{\begin{tabular}[c]{@{}c@{}}Image-level\\ (AUROC, AP)\end{tabular}}  & Obj \&texture        & MVTec AD & (\textcolor{red}{99.0}, \textcolor{red}{99.7}) & (\textcolor{blue}{98.7}, \textcolor{blue}{99.4}) & (91.2, 96.0) \\
                                            & \multirow{3}{*}{Obj} & VisA     & (\textcolor{blue}{94.6}, \textcolor{red}{95.9}) & (\textcolor{red}{95.3}, \textcolor{blue}{95.7}) & (88.5, 90.4) \\
                                            &                      & MPDD     & (\textcolor{red}{94.1}, \textcolor{red}{96.3}) & (\textcolor{blue}{91.6}, \textcolor{blue}{93.8}) & (79.7, 84.5) \\
                                            &                      & BTAD     & (93.2, \textcolor{red}{98.6}) & (\textcolor{blue}{93.8}, 96.8) & (\textcolor{red}{95.0}, \textcolor{blue}{98.4}) \\
                                            & Texture              & DAGM     & (92.7, \textcolor{blue}{81.3}) & (\textcolor{blue}{92.9}, 79.1) & (\textcolor{red}{98.5}, \textcolor{red}{94.3}) \\ \hline
\multirow{5}{*}{\begin{tabular}[c]{@{}c@{}}Pixel-level\\ (AUROC, PRO)\end{tabular}} & Obj \&texture        & MVTec AD & (\textcolor{red}{98.1}, \textcolor{blue}{92.8}) & (\textcolor{blue}{97.8}, \textcolor{red}{93.6}) & (90.5, 87.1) \\
                                            & \multirow{3}{*}{Obj} & VisA     & (\textcolor{red}{98.5}, \textcolor{red}{92.2}) & (\textcolor{blue}{98.4}, 91.2) & (95.4, \textcolor{blue}{92.2}) \\
                                            &                      & MPDD     & (\textcolor{red}{98.8}, \textcolor{blue}{94.9}) & (\textcolor{blue}{98.4}, \textcolor{red}{95.2}) & (97.2, 92.6) \\
                                            &                      & BTAD     & (\textcolor{blue}{97.4}, 74.4) & (\textcolor{red}{97.5}, \textcolor{blue}{75.1}) & (95.6, \textcolor{red}{86.1}) \\
                                            & Texture              & DAGM     & (95.9, 87.9) & (\textcolor{blue}{96.8}, \textcolor{blue}{91.9}) & (\textcolor{red}{97.5}, \textcolor{red}{95.2}) \\ \hline
                                            \noalign{\hrule height 0.5mm}
\end{tabular}
}
\caption{Comparison of ZSAD performance between ViP$^{2}$-CLIP and SOTA full-shot methods.Results of methods with \textsuperscript{*} are copied from the AnomalyCLIP paper.The best performance is highlighted in red, and the second is highlighted in blue.}
\label{tab:table6}
\end{table*}

%% file: sec/table/table_8.tex
\begin{table}[]
\resizebox{\linewidth}{!}{
\begin{tabular}{ccccc}
\hline
\noalign{\hrule height 0.5mm}
Task                                          & Datasets & $|\mathcal{C}|$ & AnomalyCLIP & ViP$^{2}$-CLIP \\ \hline
\multirow{4}{*}{\begin{tabular}[c]{@{}c@{}}Image-level\\ (AUROC, AP, F1)\end{tabular}}  
 & HeadCT   & 1   & (94.3, 94.0, 90.5)          & (\textcolor{red}{96.5}, \textcolor{red}{96.5}, \textcolor{red}{92.8})       \\
 & BrainMRI & 1 & (94.7, 95.7, 92.5) & (\textcolor{red}{96.7}, \textcolor{red}{97.6}, \textcolor{red}{94.9}) \\
 & Brain35H & 1 & (96.5, 96.7, 92.6) & (\textcolor{red}{97.1}, \textcolor{red}{97.5}, \textcolor{red}{93.4}) \\ \cline{2-5}
 & AVERAGE  & -  & (95.2, 95.5, 91.9) & (\textcolor{red}{96.8}, \textcolor{red}{97.2}, \textcolor{red}{93.7}) \\ \hline

\multirow{6}{*}{\begin{tabular}[c]{@{}c@{}}Pixel-level\\ (AUROC, PRO, F1)\end{tabular}} 
 & ISIC     & 1 & (87.8, 75.6, 69.7) & (\textcolor{red}{91.7}, \textcolor{red}{82.9}, \textcolor{red}{75.3}) \\
 & CVC-ColonDB  & 1 & (\textcolor{red}{87.8}, 76.9, 48.5) & (87.7, \textcolor{red}{86.2}, \textcolor{red}{53.5}) \\
 & CVC-ClinicDB & 1 & (88.8, 76.1, 51.7) & (\textcolor{red}{92.2}, \textcolor{red}{84.3}, \textcolor{red}{60.8}) \\
 & Endo     & 1 & (89.0, 73.5, 58.1) & (\textcolor{red}{92.3}, \textcolor{red}{82.8}, \textcolor{red}{65.2}) \\
 & Kvasir   & 1 & (85.6, 50.0, 55.1) & (\textcolor{red}{90.5}, \textcolor{red}{56.9}, \textcolor{red}{63.4}) \\ \cline{2-5}
 & AVERAGE  & -  & (87.8, 70.4, 56.6) & (\textcolor{red}{90.9}, \textcolor{red}{78.6}, \textcolor{red}{63.6}) \\ \hline
 \noalign{\hrule height 0.5mm}
\end{tabular}
}
\caption{ZSAD performance on medical images after fine-tuning by medical-domain datasets. Best performance is shown in red.}
\label{tab:table8}
\end{table}

%% file: sec/table/table_4.tex
\begin{table}[]
\resizebox{\linewidth}{!}{
\begin{tabular}{ccccccccccccc}
\hline
\noalign{\hrule height 0.5mm}
\multirow{2}{*}{State words} & \multicolumn{2}{c}{MVTec AD}                             & \multicolumn{2}{c}{VisA}                             \\
\multicolumn{1}{l}{} & Pixel-level               & Image-level              & Pixel-level               & Image-level              \\ \hline

Good/Damage & 
(90.5, 87.1, 43.1) & 
(91.2, 96.0, 92.0) & 
(95.4, 92.2, 33.6) & 
(88.5, 90.4, 84.8) \\

Normal/Abnormal & 
(90.6, 87.5, 43.6) & 
(91.1, 95.9, 92.3) & 
(95.3, 91.9, 33.3) & 
(88.3, 90.2, 84.5) \\

Perfect/Flawed & 
(90.4, 87.2, 43.7) & 
(91.7, 96.2, 92.5) & 
(95.4, 92.0, 33.5) & 
(88.3, 90.3, 84.5) \\

Flawless/Imperfect & 
(90.4, 87.4, 43.3) & 
(91.2, 95.9, 92.2) & 
(95.4, 91.9, 33.1) & 
(87.9, 90.2, 84.3) \\ \hline

\noalign{\hrule height 0.5mm}
\end{tabular}
}
\caption{Ablation on different state words in prompts.}
\label{tab:table4}
\end{table}

%% file: sec/table/table_9.tex
\begin{table*}[ht!]
\resizebox{\linewidth}{!}{
\begin{tabular}{ccccccc}
\hline
\noalign{\hrule height 0.5mm}
Object name & CLIP & WinCLIP & APRIL-GAN & AdaCLIP & AnomalyCLIP & ViP$^{2}$-CLIP \\ \hline
Candle     & 32.7 & 87.0 & 97.8 & 98.9 & 98.8 & 99.0 \\
Capsules   & 44.8 & 80.0 & 97.5 & 98.7 & 95.0 & 98.3 \\
Cashew     & 21.8 & 84.8 & 85.8 & 92.1 & 93.8 & 91.4 \\
Chewinggum & 37.8 & 95.4 & 99.5 & 99.6 & 99.3 & 99.7 \\
Fryum      & 26.0 & 87.7 & 91.9 & 94.5 & 94.6 & 92.6 \\
Macaroni1  & 52.9 & 50.3 & 98.8 & 99.2 & 98.3 & 99.5 \\
Macaroni2  & 70.4 & 44.7 & 97.8 & 98.4 & 97.6 & 98.7 \\
Pcb1       & 61.9 & 38.6 & 92.8 & 93.7 & 94.0 & 90.1 \\
Pcb2       & 27.4 & 58.7 & 89.8 & 90.8 & 92.4 & 92.0 \\
Pcb3       & 71.7 & 76.0 & 88.2 & 88.5 & 88.4 & 92.2 \\
Pcb4       & 49.2 & 91.4 & 94.5 & 96.1 & 95.7 & 96.4 \\
Pipe\_fryum & 26.9 & 83.6 & 96.0 & 96.0 & 98.2 & 95.2 \\
\hline
Mean       & 43.6 & 73.2 & 94.2 & 95.5 & 95.5 & 95.4 \\
\noalign{\hrule height 0.5mm}
\end{tabular}
}
\caption{Fine-grained performance comparison for Pixel-level AUROC on VisA.}
\label{tab:table9}
\end{table*}

%% file: sec/table/table_10.tex
\begin{table*}[ht!]
\resizebox{\linewidth}{!}{
\begin{tabular}{ccccccc}
\hline
\noalign{\hrule height 0.5mm}
Object name & CLIP & WinCLIP & APRIL-GAN & AnomalyCLIP & AdaCLIP & ViP$^{2}$-CLIP \\ \hline
Candle     & 7.9  & 77.7 & 92.3 & 96.5 & 62.2 & 96.7 \\
Capsules   & 12.0 & 39.4 & 86.1 & 78.9 & 38.3 & 93.5 \\
Cashew     & 0.3  & 78.4 & 91.5 & 91.9 & 57.5 & 94.9 \\
Chewinggum & 9.1  & 69.6 & 87.5 & 90.9 & 55.8 & 96.8 \\
Fryum      & 6.1  & 74.4 & 89.4 & 86.9 & 52.7 & 94.8 \\
Macaroni1  & 25.9 & 24.8 & 93.0 & 89.8 & 65.6 & 96.4 \\
Macaroni2  & 32.6 & 8.0  & 82.0 & 84.0 & 62.9 & 89.5 \\
Pcb1       & 18.4 & 20.7 & 87.3 & 80.7 & 47.7 & 91.4 \\
Pcb2       & 8.2  & 20.7 & 75.4 & 78.9 & 50.2 & 80.0 \\
Pcb3       & 27.2 & 43.8 & 77.2 & 76.8 & 41.8 & 84.3 \\
Pcb4       & 11.6 & 74.5 & 86.6 & 89.4 & 63.9 & 90.7 \\
Pipe\_fryum & 9.1  & 80.3 & 90.9 & 96.2 & 83.2 & 97.2 \\
\hline
Mean       & 14.0 & 51.0 & 86.6 & 86.7 & 56.8 & 92.2 \\
\noalign{\hrule height 0.5mm}
\end{tabular}
}
\caption{Fine-grained performance comparison for Pixel-level AUPRO on VisA.}
\label{tab:table10}
\end{table*}

%% file: sec/table/table_11.tex
\begin{table*}[ht!]
\resizebox{\linewidth}{!}{
\begin{tabular}{ccccccc}
\hline
\noalign{\hrule height 0.5mm}
Object name & CLIP & WinCLIP & APRIL-GAN & AnomalyCLIP & AdaCLIP & ViP$^{2}$-CLIP \\ \hline
Candle     & 0.3  & 8.9  & 39.4 & 37.8 & 42.1 & 42.4 \\
Capsules   & 1.1  & 4.2  & 49.1 & 37.8 & 54.1 & 50.7 \\
Cashew     & 2.2  & 9.6  & 22.7 & 25.8 & 35.1 & 23.8 \\
Chewinggum & 1.1  & 31.6 & 78.5 & 61.0 & 78.2 & 74.4 \\
Fryum      & 4.5  & 16.2 & 29.5 & 30.3 & 32.7 & 25.6 \\
Macaroni1  & 0.1  & 0.1  & 35.3 & 23.7 & 32.7 & 38.9 \\
Macaroni2  & 0.1  & 0.1  & 13.9 & 5.1  & 14.1 & 9.7 \\
Pcb1       & 2.3  & 0.9  & 12.2 & 12.7 & 13.7 & 13.4 \\
Pcb2       & 0.4  & 1.5  & 23.3 & 15.8 & 32.5 & 23.4 \\
Pcb3       & 1.2  & 2.1  & 21.9 & 9.3  & 35.8 & 25.7 \\
Pcb4       & 1.5  & 24.6 & 31.0 & 34.7 & 43.0 & 39.8 \\
Pipe\_fryum & 2.5  & 8.3  & 30.4 & 45.5 & 29.5 & 35.9 \\
\hline
Mean       & 1.5  & 9.0  & 32.3 & 28.3 & 37.0 & 33.6 \\
\noalign{\hrule height 0.5mm}
\end{tabular}
}
\caption{Fine-grained performance comparison for Pixel-level F1 on VisA.}
\label{tab:table11}
\end{table*}

%% file: sec/table/table_12.tex
\begin{table*}[ht!]
\resizebox{\linewidth}{!}{
\begin{tabular}{ccccccc}
\hline
\noalign{\hrule height 0.5mm}
Object name & CLIP & WinCLIP & APRIL-GAN & AnomalyCLIP & AdaCLIP & ViP$^{2}$-CLIP \\ \hline
Candle     & 55.2 & 94.8 & 82.6 & 80.9 & 96.0 & 90.4 \\
Capsules   & 61.9 & 79.4 & 62.3 & 82.8 & 87.0 & 93.8 \\
Cashew     & 68.3 & 91.2 & 86.6 & 76.0 & 88.1 & 96.4 \\
Chewinggum & 59.4 & 95.5 & 96.4 & 97.2 & 93.9 & 97.9 \\
Fryum      & 48.4 & 73.6 & 93.8 & 92.7 & 91.8 & 88.5 \\
Macaroni1  & 61.5 & 79.0 & 69.3 & 86.7 & 84.2 & 86.3 \\
Macaroni2  & 52.3 & 67.1 & 65.7 & 72.2 & 67.5 & 70.8 \\
Pcb1       & 66.1 & 72.1 & 51.0 & 85.2 & 89.2 & 90.8 \\
Pcb2       & 64.4 & 47.0 & 71.4 & 62.0 & 84.3 & 78.8 \\
Pcb3       & 50.7 & 63.9 & 66.9 & 61.7 & 78.9 & 78.3 \\
Pcb4       & 64.8 & 74.2 & 94.7 & 93.9 & 95.3 & 97.5 \\
Pipe\_fryum & 68.9 & 67.9 & 89.2 & 92.3 & 90.3 & 92.4 \\
\hline
Mean       & 60.2 & 75.5 & 77.5 & 82.0 & 87.2 & 88.5 \\
\noalign{\hrule height 0.5mm}
\end{tabular}
}
\caption{Fine-grained performance comparison for Image-level AUROC on VisA.}
\label{tab:table12}
\end{table*}

%% file: sec/table/table_13.tex
\begin{table*}[ht!]
\resizebox{\linewidth}{!}{
\begin{tabular}{ccccccc}
\hline
\noalign{\hrule height 0.5mm}
Object name & CLIP & WinCLIP & APRIL-GAN & AnomalyCLIP & AdaCLIP & ViP$^{2}$-CLIP \\ \hline
Candle     & 56.1 & 95.4 & 86.0 & 82.6 & 96.6 & 92.6 \\
Capsules   & 74.9 & 87.9 & 74.5 & 89.4 & 92.5 & 96.7 \\
Cashew     & 80.9 & 96.0 & 93.8 & 89.3 & 94.9 & 98.4 \\
Chewinggum & 77.3 & 98.2 & 98.4 & 98.8 & 97.5 & 99.1 \\
Fryum      & 70.4 & 86.9 & 97.1 & 96.6 & 96.3 & 94.5 \\
Macaroni1  & 60.3 & 80.0 & 67.4 & 85.5 & 83.1 & 87.9 \\
Macaroni2  & 49.9 & 65.1 & 64.8 & 70.8 & 67.7 & 71.1 \\
Pcb1       & 69.9 & 73.0 & 55.3 & 86.7 & 89.4 & 91.0 \\
Pcb2       & 62.5 & 46.1 & 73.4 & 64.4 & 85.6 & 78.9 \\
Pcb3       & 50.6 & 63.1 & 70.4 & 69.4 & 82.4 & 81.6 \\
Pcb4       & 59.6 & 70.1 & 94.8 & 94.3 & 95.5 & 96.8 \\
Pipe\_fryum & 82.1 & 82.1 & 94.5 & 96.3 & 95.1 & 96.3 \\
\hline
Mean       & 66.2 & 78.7 & 80.9 & 85.3 & 89.7 & 90.4 \\
\noalign{\hrule height 0.5mm}
\end{tabular}
}
\caption{Fine-grained performance comparison for Image-level AP on VisA.}
\label{tab:table13}
\end{table*}

%% file: sec/table/table_14.tex
\begin{table*}[ht!]
\resizebox{\linewidth}{!}{
\begin{tabular}{ccccccc}
\hline
\noalign{\hrule height 0.5mm}
Object name & CLIP & WinCLIP & APRIL-GAN & AnomalyCLIP & AdaCLIP & ViP$^{2}$-CLIP \\ \hline
Candle     & 67.4 & 90.6 & 77.9 & 75.6 & 89.5 & 83.0 \\
Capsules   & 76.9 & 80.5 & 78.0 & 82.2 & 85.0 & 90.8 \\
Cashew     & 80.2 & 88.9 & 85.4 & 80.3 & 86.8 & 94.0 \\
Chewinggum & 80.0 & 93.8 & 93.2 & 94.8 & 91.5 & 96.4 \\
Fryum      & 80.0 & 80.0 & 91.5 & 90.1 & 88.2 & 88.2 \\
Macaroni1  & 71.5 & 74.2 & 70.8 & 80.4 & 80.4 & 79.6 \\
Macaroni2  & 66.7 & 68.8 & 69.3 & 71.2 & 68.8 & 71.3 \\
Pcb1       & 68.1 & 70.2 & 66.9 & 78.8 & 81.5 & 85.0 \\
Pcb2       & 68.4 & 67.1 & 69.1 & 67.8 & 77.5 & 74.1 \\
Pcb3       & 66.4 & 67.6 & 66.7 & 66.4 & 73.6 & 73.3 \\
Pcb4       & 69.6 & 75.7 & 87.3 & 87.8 & 89.7 & 92.4 \\
Pipe\_fryum & 80.8 & 80.3 & 88.1 & 89.8 & 89.0 & 89.5 \\
\hline
Mean       & 73.0 & 78.2 & 78.7 & 80.4 & 83.5 & 84.8 \\
\noalign{\hrule height 0.5mm}
\end{tabular}
}
\caption{Fine-grained performance comparison for Image-level F1 on VisA.}
\label{tab:table14}
\end{table*}

%% file: sec/table/table_15.tex
\begin{table*}[ht!]
\resizebox{\linewidth}{!}{
\begin{tabular}{ccccccc}
\hline
\noalign{\hrule height 0.5mm}
Object name & CLIP & WinCLIP & VAND & AnomalyCLIP & AdaCLIP & ViP$^{2}$-CLIP \\ \hline
Carpet     & 18.0 & 90.9 & 98.4 & 98.8 & 98.6 & 99.2 \\
Bottle     & 19.6 & 85.7 & 83.5 & 90.4 & 92.8 & 89.4 \\
Hazelnut   & 27.6 & 95.7 & 96.1 & 97.2 & 98.6 & 96.7 \\
Leather    & 12.7 & 95.5 & 99.1 & 98.6 & 99.3 & 99.2 \\
Cable      & 44.1 & 61.3 & 72.3 & 78.9 & 76.6 & 73.2 \\
Capsule    & 58.0 & 87.0 & 92.0 & 95.8 & 94.3 & 94.4 \\
Grid       & 11.8 & 79.4 & 95.8 & 97.3 & 91.1 & 97.7 \\
Pill       & 45.6 & 72.7 & 76.2 & 91.8 & 86.7 & 87.5 \\
Transistor & 42.2 & 83.7 & 62.4 & 70.8 & 63.5 & 64.6 \\
Metal Nut  & 33.3 & 49.3 & 65.5 & 74.6 & 68.2 & 76.8 \\
Screw      & 72.3 & 91.1 & 97.8 & 97.5 & 97.8 & 98.7 \\
Toothbrush & 26.0 & 86.2 & 95.8 & 91.9 & 97.2 & 93.4 \\
Zipper     & 52.5 & 91.7 & 91.1 & 91.3 & 95.7 & 97.0 \\
Tile       & 34.6 & 79.1 & 92.7 & 94.7 & 89.5 & 93.7 \\
Wood       & 36.1 & 85.1 & 95.8 & 96.4 & 94.0 & 95.7 \\
\hline
Mean       & 35.6 & 82.3 & 87.6 & 91.1 & 89.6 & 90.5 \\
\noalign{\hrule height 0.5mm}
\end{tabular}
}
\caption{Fine-grained performance comparison for Pixel-level AUROC on MVTec.}
\label{tab:table15}
\end{table*}

%% file: sec/table/table_16.tex
\begin{table*}[ht!]
\resizebox{\linewidth}{!}{
\begin{tabular}{ccccccc}
\hline
\noalign{\hrule height 0.5mm}
Object name & CLIP & WinCLIP & VAND & AnomalyCLIP & AdaCLIP & ViP$^{2}$-CLIP \\ \hline
Carpet     & 6.2  & 66.3 & 48.5 & 90.0  & 38.1 & 97.9 \\
Bottle     & 0.3  & 69.9 & 45.6 & 80.8  & 39.0 & 83.5 \\
Hazelnut   & 4.7  & 81.4 & 70.3 & 92.5  & 19.4 & 93.1 \\
Leather    & 1.4  & 86.0 & 72.4 & 92.2  & 57.0 & 98.8 \\
Cable      & 8.8  & 39.4 & 25.7 & 64.0  & 43.1 & 66.5 \\
Capsule    & 31.6 & 63.7 & 51.3 & 87.6  & 60.3 & 92.7 \\
Grid       & 0.2  & 49.3 & 31.6 & 75.4  & 57.9 & 89.5 \\
Pill       & 5.3  & 66.9 & 65.4 & 88.1  & 40.5 & 93.1 \\
Transistor & 8.6  & 45.5 & 21.3 & 58.2  & 27.1 & 55.1 \\
Metal Nut  & 0.9  & 39.6 & 38.4 & 71.1  & 63.9 & 79.8 \\
Screw      & 48.5 & 70.2 & 67.1 & 88.0  & 16.1 & 94.2 \\
Toothbrush & 2.3  & 67.9 & 54.5 & 88.5  & 58.9 & 88.8 \\
Zipper     & 20.8 & 72.1 & 10.7 & 65.4  & 18.1 & 89.1 \\
Tile       & 6.3  & 54.5 & 26.7 & 87.4  & 25.7 & 89.2 \\
Wood       & 12.9 & 56.3 & 31.1 & 91.5  & 2.4 & 95.5 \\
\hline
Mean       & 10.6 & 61.9 & 44.0 & 81.4  & 37.8 & 87.1 \\
\noalign{\hrule height 0.5mm}
\end{tabular}
}
\caption{Fine-grained performance comparison for Pixel-level AUPRO on MVTec.}
\label{tab:table16}
\end{table*}

%% file: sec/table/table_17.tex
\begin{table*}[ht!]
\resizebox{\linewidth}{!}{
\begin{tabular}{ccccccc}
\hline
\noalign{\hrule height 0.5mm}
Object name & CLIP & WinCLIP & VAND & AnomalyCLIP & AdaCLIP & ViP$^{2}$-CLIP \\ \hline
Carpet     & 3.2  & 33.9 & 65.7 & 57.0 & 59.5 & 65.5 \\
Bottle     & 10.9 & 49.4 & 53.4 & 51.6 & 32.2 & 52.3 \\
Hazelnut   & 4.2  & 39.1 & 50.5 & 47.6 & 34.2 & 49.9 \\
Leather    & 1.3  & 30.8 & 50.0 & 33.2 & 66.6 & 43.0 \\
Cable      & 5.6  & 12.2 & 23.9 & 18.9 & 37.0 & 24.1 \\
Capsule    & 4.8  & 14.3 & 33.1 & 31.0 & 63.3 & 34.4 \\
Grid       & 1.4  & 13.7 & 40.8 & 32.0 & 49.8 & 41.3 \\
Pill       & 7.4  & 11.8 & 27.7 & 35.5 & 31.2 & 26.0 \\
Transistor & 9.2  & 27.0 & 19.0 & 18.8 & 31.7 & 16.0 \\
Metal Nut  & 21.0 & 23.8 & 28.0 & 33.1 & 28.6 & 35.1 \\
Screw      & 6.4  & 11.3 & 41.7 & 33.4 & 67.4 & 46.4 \\
Toothbrush & 3.0  & 10.5 & 48.1 & 29.0 & 47.8 & 31.9 \\
Zipper     & 5.2  & 27.8 & 40.5 & 45.0 & 18.0  & 53.2 \\
Tile       & 13.2 & 30.8 & 66.5 & 64.9 & 52.9 & 68.0 \\
Wood       & 7.4  & 35.4 & 60.3 & 55.2 & 55.6 & 58.7 \\
\hline
Mean       & 6.9  & 24.8 & 43.3 & 39.1 & 45.1 & 43.1 \\
\noalign{\hrule height 0.5mm}
\end{tabular}
}
\caption{Fine-grained performance comparison for Pixel-level F1 on MVTec.}
\label{tab:table17}
\end{table*}

%% file: sec/table/table_18.tex
\begin{table*}[ht!]
\resizebox{\linewidth}{!}{
\begin{tabular}{ccccccc}
\hline
\noalign{\hrule height 0.5mm}
Object name & CLIP & WinCLIP & VAND & AnomalyCLIP & AdaCLIP & ViP$^{2}$-CLIP \\ \hline
Carpet & 86.9 & 99.3 & 99.4 & 100 & 99.9 & 99.9 \\
Bottle & 20.5 & 98.6 & 91.9 & 88.7 & 97.9 & 93.7 \\
Hazelnut & 53 & 92.3 & 89.4 & 97.2 & 97.2 & 93.5 \\
Leather & 98.5 & 100 & 99.7 & 99.8 & 99.9 & 100 \\
Cable & 51.4 & 85 & 88.1 & 70.3 & 65.5 & 75.3 \\
Capsule & 60.7 & 68.6 & 79.9 & 89.5 & 86.5 & 90.2 \\
Grid & 77 & 99.2 & 86.5 & 97.8 & 99.4 & 99.7 \\
Pill & 61.2 & 81.5 & 80.9 & 81.1 & 84.6 & 84.6 \\
Transistor & 45.6 & 89.1 & 80.8 & 93.9 & 82.2 & 79.9 \\
Metal Nut & 68.2 & 96.2 & 68.4 & 92.4 & 75.9 & 77.3 \\
Screw & 62.2 & 71.6 & 85.1 & 82.1 & 97.8 & 90 \\
Toothbrush & 23.6 & 85.3 & 53.6 & 85.3 & 93.3 & 94.2 \\
Zipper & 90 & 91.2 & 89.6 & 98.4 & 96.4 & 92.9 \\
Tile & 98.9 & 99.9 & 99.9 & 100 & 99.9 & 98.5 \\
Wood & 99.6 & 97.6 & 99 & 96.9 & 99.0 & 98.6 \\
\hline
Mean & 66.5 & 90.4 & 86.1 & 91.6 & 90.1 & 91.2 \\
\noalign{\hrule height 0.5mm}
\end{tabular}
}
\caption{Fine-grained performance comparison for Image-level AUROC on MVTec.}
\label{tab:table18}
\end{table*}

%% file: sec/table/table_19.tex
\begin{table*}[ht!]
\resizebox{\linewidth}{!}{
\begin{tabular}{ccccccc}
\hline
\noalign{\hrule height 0.5mm}
Object name & CLIP & WinCLIP & VAND & AnomalyCLIP & AdaCLIP & ViP$^{2}$-CLIP \\ \hline
Carpet & 95.6 & 97.8 & 99.8 & 100 & 99.9 & 100 \\
Bottle & 65 & 97.6 & 97.6 & 96.8 & 99.4 & 98.1 \\
Hazelnut & 67.2 & 88.6 & 94.6 & 98.5 & 98.4 & 96.8 \\
Leather & 99.5 & 100 & 99.9 & 99.9 & 99.9 & 100 \\
Cable & 66.2 & 89.8 & 92.9 & 81.7 & 81.8 & 84.5 \\
Capsule & 88.5 & 90.5 & 95.4 & 97.8 & 97.1 & 97.9 \\
Grid & 90.6 & 98.2 & 94.9 & 99.3 & 99.8 & 99.9 \\
Pill & 89.9 & 96.4 & 96.1 & 95.3 & 96.6 & 96.6 \\
Transistor & 46.3 & 84.9 & 77.5 & 92.1 & 82.7 & 80 \\
Metal Nut & 91.1 & 99.1 & 91.8 & 98.2 & 94.5 & 94.7 \\
Screw & 81.6 & 87.7 & 93.6 & 92.9 & 96.5 & 96.5 \\
Toothbrush & 60.7 & 94.5 & 71.6 & 93.9 & 97.7 & 97.5 \\
Zipper & 96.6 & 97.5 & 97.1 & 99.5 & 99.1 & 98.2 \\
Tile & 99.5 & 100 & 100 & 100 & 100 & 99.5 \\
Wood & 99.9 & 99.3 & 99.7 & 99.2 & 99.7 & 99.6 \\
\hline
Mean & 82.6 & 95.6 & 93.5 & 96.4 & 95.6 & 96 \\
\noalign{\hrule height 0.5mm}
\end{tabular}
}
\caption{Fine-grained performance comparison for Image-level AP on MVTec.}
\label{tab:table19}
\end{table*}

%% file: sec/table/table_20.tex
\begin{table*}[ht!]
\resizebox{\linewidth}{!}{
\begin{tabular}{ccccccc}
\hline
\noalign{\hrule height 0.5mm}
Object name & CLIP & WinCLIP & VAND & AnomalyCLIP & AdaCLIP & ViP$^{2}$-CLIP \\ \hline
Carpet & 89.4 & 97.8 & 98.3 & 99.4 & 99.4 & 99.4 \\
Bottle & 86.3 & 97.6 & 92.1 & 90.9 & 92.4 & 92.4 \\
Hazelnut & 77.8 & 88.6 & 87 & 92.6 & 91.2 & 91.2 \\
Leather & 96.7 & 100 & 98.9 & 99.5 & 99.5 & 100 \\
Cable & 76 & 84.8 & 84.5 & 77.4 & 76.0 & 80.2 \\
Capsule & 90.5 & 93.5 & 91.5 & 91.7 & 93.1 & 93 \\
Grid & 88.4 & 98.2 & 89.1 & 97.3 & 98.3 & 99.1 \\
Pill & 91.6 & 91.6 & 91.6 & 92.1 & 91.8 & 91.8 \\
Transistor & 57.1 & 80 & 73.9 & 83.7 & 75.9 & 71.8 \\
Metal Nut & 89.4 & 95.3 & 89.4 & 93.7 & 89.4 & 90.3 \\
Screw & 85.6 & 85.9 & 89.3 & 88.3 & 91.3 & 91.3 \\
Toothbrush & 83.3 & 88.9 & 83.3 & 90 & 93.5 & 93.5 \\
Zipper & 91.5 & 93.4 & 90.8 & 97.9 & 95.1 & 92.4 \\
Tile & 97 & 99.4 & 99.4 & 100 & 99.4 & 96.4 \\
Wood & 99.2 & 95.2 & 97.4 & 96.6 & 96.7 & 96.8 \\
\hline
Mean & 86.7 & 92.7 & 90.4 & 92.7 & 92.3 & 92 \\
\noalign{\hrule height 0.5mm}
\end{tabular}
}
\caption{Fine-grained performance comparison for Image-level F1 on MVTec.}
\label{tab:table20}
\end{table*}